\RequirePackage{fix-cm}
\RequirePackage{amsmath}
\RequirePackage{amssymb}
\documentclass[smallextended]{svjour3}       
\smartqed  
\usepackage{graphicx}
\usepackage{xcolor}

\newenvironment{rev}{}{} 

\newcommand{\revisions}[1]{\begin{rev}#1\end{rev}}

\newcommand{\x}{\mathbf{x}}
\newcommand{\y}{\mathbf{y}}
\newcommand{\s}{\mathbf{s}}

\journalname{Machine Learning}

\begin{document}

\title{Learning From Positive and Unlabeled Data: \\A Survey
}
\author{Jessa Bekker         \and
        Jesse Davis 
}

\institute{J. Bekker \at
              KU Leuven, Leuven, Belgium \\
              \email{jessa.bekker@kuleuven.be}          
           \and
           J. Davis \at
              KU Leuven, Leuven, Belgium \\
              \email{jesse.davis@kuleuven.be}
}

\date{Received: 12 November 2018 / Accepted: 1 March 2020}

\maketitle

\begin{abstract}
Learning from positive and unlabeled data or PU learning is the setting where a learner only has access to positive examples and unlabeled data. The assumption is that the unlabeled data can contain both positive and negative examples. This setting has attracted increasing interest within the machine learning literature as this type of data naturally arises in applications such as medical diagnosis and knowledge base completion. This article provides a survey of the current state of the art in PU learning. It proposes seven key research questions that commonly arise in this field and provides a broad overview of how the field has tried to address them. 
\keywords{Classification \and Weakly supervised learning \and PU learning}
\subclass{68T05}
\end{abstract}

\section{Introduction}

The goal of binary classification is to learn a model that is able to distinguish between positive and negative examples. To do so, an algorithm has access to training data. In the most traditional setting, this data contains both positive and negative examples and is fully labeled, that is, the class value is not missing for any training example. This is among the most widely studied problems in machine learning.

Learning from positive and unlabeled data or PU learning is a variant of this classical set up where the training data consists of positive and unlabeled examples. The assumption is that each unlabeled example could belong to either the positive or negative class. The term PU learning first began to appear in the early 2000s and there has been a surge of interest in this setting in recent years~\cite{liu2003building,denis2005learning,Li2005LearningFP,elkan2008learning,mordelet2014bagging,Plessis2015ConvexFF}. It fits within the long standing interest in developing learning algorithms that do not require fully supervised data, such as learning from positive-only or one-class data~\cite{khan_madden_2014} and semi-supervised learning~\cite{chapelle2009semi}. PU learning differs from the former in that it explicitly incorporates unlabeled data into the learning process. It is related to the latter in that it specializes the standard semi-supervised setting, where typically some labeled examples for all classes are available.

One reason that PU learning has attracted attention is that PU data naturally arises in many significant applications. The following are three illustrative examples of applications characterized by PU data. First, personalized advertising uses visited pages and clicks as positive examples of pages and ads of interest. However, all other pages or ads are not necessarily uninteresting and should therefore not be treated as negative examples but as unlabeled ones. Second, medical records usually only list which diseases a patient has been diagnosed with and they usually do not include which diseases a patient does not have. However, the absence of a diagnosis does not mean that a patient does not have a disease. A patient may simply elect not to go to a doctor and moreover many diseases, such as diabetes, often go undiagnosed~\cite{claesen2015building}. Third, consider the task of knowledge base (KB) completion where the goal is to predict which other tuples should belong in an automatically constructed KB. Here, the training data consists of the tuples already in the KB. However, KBs typically only contain facts (i.e., true statements), so there are no negative examples and the truth value of any tuple not in the KB should be considered unknown~\cite{Galarraga2015,zupanc2018estimating}.

Motivated by these significant applications, researchers have taken a keen interest in analyzing the PU learning setting. Within PU learning, people have addressed a number of different tasks using a variety of techniques. Despite the breadth, at a high level, the key research questions about PU learning can be formulated rather straightforwardly as:
\begin{enumerate}
    \item How can we formalize the problem of learning from PU data?
    \item What assumptions are typically made about PU data in order to facilitate the design of learning algorithms? 
    \item Can we estimate the class prior from PU data and why is this useful?
    \item How can we learn a model from PU data?
    \item How can we evaluate models in a PU setting?
    \item When and why does PU data arise in real-world applications?
    \item How does PU learning relate to other areas of machine learning?
\end{enumerate}
This survey is structured around giving a comprehensive overview about how the PU  learning research community is tackling each of these questions. It concludes with some perspectives about future directions for PU learning research. 

\section{Preliminaries on PU Learning}
\label{sec:prelim}
Learning from positive and unlabeled data (PU learning) is a special case of binary classification. Therefore, we first review binary classification before formally describing the PU learning setting. Then we introduce the labeling mechanism, which is a key concept in PU learning. Finally, we distinguish between two PU learning settings: the single-training-set and case-control scenarios.

\subsection{Binary Classification}

The goal of binary classification is to train a classifier that can distinguish between two classes of instances, based on their attributes. By convention, the two classes are called ``positive'' and ``negative''.  To train a binary classifier, the machine learning algorithm has access to a set of training examples. Each training example is a tuple $(x,y)$, where $x$ is the vector of attribute values and $y$ is the class value. An example is positive if $y=1$ and negative if $y=0$. Traditional learning algorithms work in a supervised setting, where the training data is assumed to be fully labeled. That is, the class value for each training example is observed. Table~\ref{tab:train} shows an example of a fully labeled training set. To enable training a correct classifier, the training data is assumed to be an independent and identically distributed (i.i.d.) sample of the real distribution:
\begin{align}
\x &\sim f(x) \nonumber\\
&\sim \alpha f_+(x)+(1-\alpha)f_-(x) \label{eq:distr},
\end{align}
with class prior $\alpha=\Pr(y=1)$ and probability density functions of the true distribution $f$ and the positive and negative examples $f_+$ and $f_-$ respectively.

\begin{table}[t]
	\centering
	\caption[Labeled training set example]{\textbf{Labeled training set example.} The vector of attribute values are the first 5 rows: $x=[$age, diabetes family, fatigue, pee/day, blurred vision$]$}
	\label{tab:train}
	\begin{tabular}{ccccc|c}
		age & diabetes family & fatigue & pee/day & blurred vision & $y$\\\hline
		25 & yes & yes & 7 & no & 0 \\
		63 & no & yes & 10 & no & 1 \\
		49 & no & no & 4 & no & 0 \\
		34 & no & yes & 6 & yes & 1
	\end{tabular}
\end{table}

\subsection{PU Learning}
The goal of PU learning is the same as general binary classification: train a classifier that can distinguish between positive and negative examples based on the attributes. However, during the learning phase, only some of the positive examples in the training data are labeled and none of the negative examples are. 

\begin{table}[t]
	\centering
	\caption[Positive and Unlabeled training set example]{\textbf{Positive and Unlabeled training set example} for the same dataset as the on in Table~\ref{tab:train}.}
	\label{tab:trainpu}
	\begin{tabular}{ccccc|cc}
		age & diabetes family & fatigue & pee/day & blurred vision & $y$ & $s$\\\hline
		25 & yes & yes & 7 & no & ? & 0\\
		63 & no & yes & 10 & no & 1  & 1\\
		49 & no & no & 4 & no & ?  & 0\\
		34 & no & yes & 6 & yes & ? & 0
	\end{tabular}
\end{table}

We represent a PU dataset as a set of triplets $(x,y,s)$ with $x$  a vector of attributes, $y$ the class and $s$ a binary variable representing whether the tuple was selected to be labeled. The class $y$ is not observed, but information about it can be derived from the value of $s$. If the example is labeled $s=1$, then it belongs to the positive class: $\Pr(y=1|s=1)=1$. When the example is unlabeled $s=0$, then it can belong to either class. Table~\ref{tab:trainpu} gives an example of a positive and unlabeled version of a training set. Table~\ref{tab:notation} gives an overview of the notation used in this article.

\begin{table}[t]
	\centering
	\caption[Notation]{\textbf{Notation} used in this article.}
	\label{tab:notation}
	\begin{tabular}{cl}
		Symbol    & Description \\\hline
		$x$	&	The vector of attributes of an example	\\
		$\x$	&	A set of vectors of attributes of examples	\\
		$y$	&	Indicator variable for an example to be positive	\\
		$\y$	&	A set of indicator variables for examples to be positive	\\
		$s$	&	Indicator variable for an example to be labeled	\\
		$\s$	&	A set of indicator variables for examples to be labeled	\\
		$\alpha$	&	Class prior $\alpha=\Pr(y=1)$	\\
		$c$	&	Label frequency $c=\Pr(s=1 \vert y=1)$	\\
		$e$	&	Propensity score function $e(x)=\Pr(s=1 \vert y=1, x)$	\\
		$f(x)$ & Probability density function of the instance space (true population)\\
		$f_+(x)$ & Probability density function of the positive instance space\\
		$f_-(x)$ & Probability density function of the negative instance space\\
		$f_l(x)$ & probability density function of the labeled instance space\\
		$f_u(x)$ & Probability density function of the unlabeled instance space\\
		$\hat  \bullet$	&	An estimate for $\bullet$.\\
	\end{tabular}
\end{table}

\subsection{Labeling Mechanism}

The labeled positive examples are selected from the complete set of positive examples according to a probabilistic labeling mechanism, where each positive example $x$  has the probability $e(x) = \Pr(s=1|y=1,x)$ of being selected to be labeled, called the \emph{propensity score}~\cite{bekker2019ecml}. Hence, the labeled distribution is a biased version of the positive distribution:
\begin{align}
    f_l(x) = \frac{e(x)}{c}f_+(x), \label{eq:labdistr}
\end{align}
with $f_l(x)$ and $f_+(x)$ the probability density functions of the labeled and positive distributions respectively. The normalization constant $c$ is the \emph{label frequency}, which is the fraction of positive examples that are labeled $c\revisions{=\mathbb{E}_x[e(x)]}=\Pr(s=1|y=1)$. This can be seen from the following derivation:
\begin{align*}
f_l(x)&= \Pr(x|s=1)\\
&= \Pr(x|s=1,y=1)&& \textit{\#by PU definition} \\
&= \frac{\Pr(s=1|x,y=1)}{\Pr(s=1|y=1)}\Pr(x|y=1)&& \textit{\#Bayes' rule} \\
&= \frac{e(x)}{c}f_+(x)
\end{align*}

\subsection{The Single-Training-Set and Case-Control Scenarios}
The positive and unlabeled examples in PU data can originate from two scenarios. Either they come from a single training set, or they come from two independently drawn datasets, one with all positive examples and one with all unlabeled examples. These scenarios are called the single-training-set scenario and the case-control scenario respectively.

The \emph{single-training-set scenario} assumes that the positive and unlabeled data examples come from the same dataset and that this dataset is an i.i.d. sample from the real distribution, like for supervised classification. A fraction $c$ from the positive examples are selected to be labeled\revisions{, following their individual propensity scores e(x)}, therefore, the dataset has a fraction $\alpha c$ of labeled examples.
\begin{align}
\x &\sim f(x) \nonumber \\
&\sim \alpha f_+(x)+(1-\alpha) f_-(x) \nonumber \\
 &\sim \alpha \revisions{c} f_l(x) + (1-\alpha \revisions{c}) f_u(x). \label{eq:sts_tot}
\end{align}
\revisions{
This scenario arises, for example, in personalized advertising, where users only click a subset of the ads of interest. It can also occur in survey data that suffers from under-reporting. That is, sometimes respondents purposely provide incorrect negative responses such as falsely denying that you are a smoker.
}

The \emph{case-control scenario} assumes that the positive and unlabeled examples come from two independent datasets and that the unlabeled dataset is an i.i.d. sample from the real distribution:
\begin{align}
\x|\s=\mathbf{0} &\sim f_u(x) \nonumber \\
&\sim f(x) \nonumber\\
&\sim\alpha f_+(x)+(1-\alpha) f_-(x). \label{eq:cc_unl}
\end{align}

\revisions{
\noindent This scenario comes from the setting where two datasets are
used and one is known to only have positive examples. For example, when trying to predict
one’s socioeconomic status from health record, positive examples could
be gathered from health centers in upper-class neighborhoods and unlabeled
examples from a random selection of health centers.
}

The observed positive examples are generated from the same distribution in both the single-training-set and case-control scenario. Hence, in both scenarios the learner has access to a set of examples drawn i.i.d. from the true distribution and a set of examples that are drawn from the positive distribution according to the labeling mechanism that is defined by the propensity score $e(x)$. As a result, most methods can handle both scenarios, but the derivation differs. Consequently, one must always consider the scenario when interpreting results and using software.

The single-training-set scenario has received substantially more attention in the literature. Therefore, this survey  assumes this scenario. When methods that were originally proposed in a case-control scenario are discussed on a level where this distinction is necessary, we either convert them to the single-training-set scenario or explicitly state that the case-control scenario is assumed.

\subsection{Relationship Between the Class Prior and the Label Frequency}
\label{sec:relalphac}
The class prior $\alpha$ and the label frequency $c$  are closely related to each other. Given a PU dataset, if one is known, the expected value of the other can be calculated. The label frequency is defined as the fraction of positive examples that are labeled in all the data:
\begin{align*}
    c &= \Pr(s=1|y=1)\\
      &= \frac{\Pr(s=1,y=1)}{\Pr(y=1)}\\
      &= \frac{\Pr(s=1)}{\Pr(y=1)}. &\#\textit{by PU definition}
\end{align*}
The probability $\Pr(s=1)$ can be counted in the data as the fraction of labeled examples. The probability $\Pr(y=1)$ is related to the class prior. In the single-training-set scenario, it is equal to the class prior. However, in the case\revisions{-}control scenario, the class prior is defined in the unlabeled data: $\alpha=\Pr(y=1|s=0)$. Here, the probability $\Pr(y=1)$ is the following:
\begin{align*}
\Pr(y=1)&=\Pr(y=1|s=0)\Pr(s=0)+\Pr(y=1|s=1)\Pr(s=1)\\
&= \alpha\Pr(s=0)+\Pr(s=1).
\end{align*}
To summarize, the conversions between $c$ and $\alpha$ are done as follows:
\begin{align}
c&=\frac{\Pr(s=1)}{\alpha}	&&\textit{\# single-training-set scenario} \label{eq:calpha}\\
c&=\frac{\Pr(s=1)}{\alpha\left(1-\Pr(s=1)\right)+\Pr(s=1)} 	&&\textit{\# case-control scenario}\label{eq:calpha_cc}\\
\alpha&=\frac{1-c}{c}\frac{\Pr(s=1)}{1-\Pr(s=1)}. &&\textit{\# case-control scenario} \label{eq:alphac_cc}
\end{align}

\section{Assumptions to Enable PU Learning}
\label{sec:assumptions}

Learning from PU data is not straightforward. There are two possibilities to explain why an example is unlabeled, either:
\begin{enumerate}
    \item It is truly a negative example; or
    \item It is a positive example, but simply was not selected by the labeling mechanism to have its label observed.
\end{enumerate} Therefore, in order to enable learning with positive and unlabeled data, it is necessary to make assumptions about either the labeling mechanism, the class distributions in the data, or both. The class prior plays an important role in PU learning and many PU learning methods require it as an input. To enable estimating it directly from PU data, additional assumptions need to be made. This section discusses the most commonly made labeling mechanism and data assumptions to enable PU learning as well as the assumptions made to enable estimating the class prior from PU data.

\subsection{Label Mechanism Assumptions}
One approach is to make assumptions about the labeling mechanism. That is, how the examples with an observed positive label were selected.

\subsubsection{Selected Completely At Random}
\begin{figure}[t]
	\centering
	\includegraphics[scale=0.35]{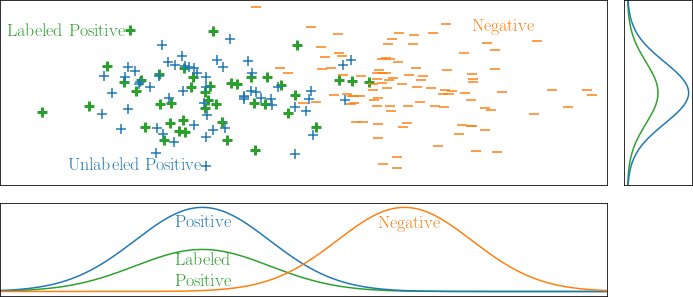}
	\caption[Example of SCAR PU data]{\textbf{Example of SCAR PU data.} The labeled examples are selected uniformly at random from the positive examples.}
	\label{fig:scar}
	\vspace{25pt}
	\includegraphics[scale=0.35]{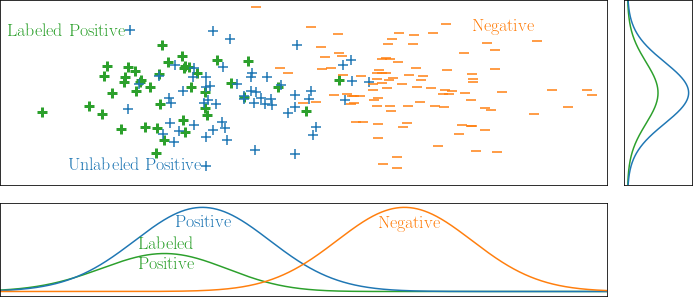}
	\caption[Example of SAR PU and PGPU data]{\textbf{Example of SAR PU and PGPU data.} The labeled examples are a biased sample of the positive examples. The larger the probabilistic gap, the more likely a positive example is selected to be labeled. This means that  positive examples which resemble negative examples more, are less likely to be labeled}
	\label{fig:sar1}
	\vspace{25pt}
	\includegraphics[scale=0.35]{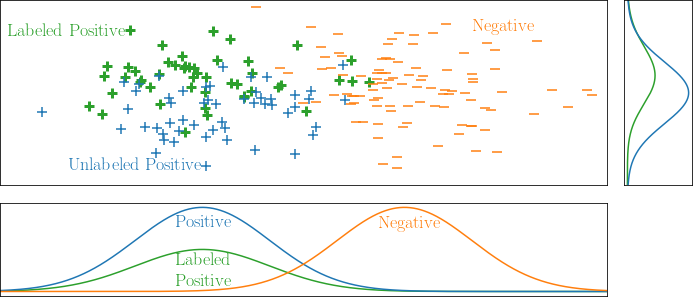}
	\caption[Example of SAR PU data]{\textbf{Example of SAR PU data.}  The labeled examples are a biased sample of the positive examples. In this case, the labeling mechanism is independent of the probabilistic gap.}
	\label{fig:sar2}
\end{figure}

The Selected Completely At Random (SCAR) assumption lies at the basis of most PU learning methods, for example, biased learning methods (Section~\ref{sec:penlearning}) and methods that directly incorporate the class prior (Section~\ref{sec:classpriorincorp}). It assumes that the set of labeled examples is a uniform subset of the set of positive examples~\cite{elkan2008learning}. Figure~\ref{fig:scar} shows an examples of a PU dataset under the SCAR assumption. \revisions{This assumption is motivated by the case-control scenario, where it is often reasonable to assume that the labeled dataset is an i.i.d. sample from the positive distribution. However, the SCAR assumption owes its popularity to its ability to reduce PU learning to standard binary classification. This enables applying standard learners to PU problems by either making minor modifications to the data (e.g., weighting it) or the underlying learning algorithm.}

\begin{definition}[Selected Completely At Random (SCAR)] Labeled examples are selected completely at random, independent from their attributes, from the positive distribution. The propensity score $e(x)$, which is the probability for selecting a positive example is constant and equal to the \emph{label frequency} $c$:
	\begin{align*}
	e(x) = \Pr(s=1|x,y=1)=\Pr(s=1|y=1)=c.
	\end{align*}
\end{definition}

Under this assumption, the set of labeled examples is an i.i.d. sample from the positive distribution. Indeed, Equation~\ref{eq:labdistr} simplifies to $f_l(x)=f_+(x)$.

Under the SCAR assumption, the probability for an example to be labeled is directly proportional to the probability for an example to be positive:
\begin{align*}
\Pr(s=1|x)=c\Pr(y=1|x).
\end{align*}
This enables the use of \emph{non-traditional classifiers}, which are classifiers that predict $\Pr(s=1|x)$, which are learned by considering the unlabeled examples as negative~\cite{elkan2008learning}. These non-traditional classifiers have various interesting properties:
\begin{itemize}
	\item Non-traditional classifiers preserve the ranking order~\cite{elkan2008learning}:
	\begin{align*}
	\Pr(y=1|x_1)>\Pr(y=1|x_2)\Leftrightarrow\Pr(s=1|x_1)>\Pr(s=1|x_2).    
	\end{align*}
	\item Training a traditional classifier subject to a desired expected recall, is equivalent to training a non-traditional classifier subject to that  recall~\cite{liu2002partially,blanchard2010semi}
	\item Given the label frequency (or class prior), a probabilistic non-traditional classifier can be converted to a traditional classifier, by dividing the outputs by the label frequency $\Pr(y=1|x)=\Pr(s=1|x)/c$~\cite{elkan2008learning}.
\end{itemize}

The SCAR assumption was introduced in analogy with the \emph{Missing Completely A Random assumption (MCAR)} that is common when working with missing data~\cite{rubin1976inference,little2002statistical}. However, there is a notable difference between the two assumptions. In MCAR data, the missingness of the variable cannot depend on the value of the variable, where in PU learning this is necessarily the case because all negative labels are missing. The class values are missing completely at random only if just the population of positive examples is considered.  Moreno et al. (2012) proposed a new missingness class: \emph{Missing Completely At Random-Class Dependent (MAR-C)}, SCAR belongs to this category.

\subsubsection{Selected At Random}
The Selected At Random (SAR) assumption, is the most general assumption about the labeling mechanism: the probability for selecting positive examples to be labeled depends on its attribute values~\cite{bekker2019ecml}.
Figures~\ref{fig:sar1} and~\ref{fig:sar2} show examples of PU datasets under the SAR assumption. \revisions{This general assumption is motivated by the fact that many PU learning applications suffer from labeling bias. For example, whether someone clicks on a sponsored search ad is influenced by the position in which it is placed. Similarly, whether a patient suffering from a disease will visit a doctor depends on her socioeconomic status and the severity of her symptoms. }

\begin{definition}[Selected At Random (SAR)] Labeled examples are a biased sample from the positive distribution, where the bias completely depends on the attributes and is defined by the \emph{propensity score} $e(x)$:
	\begin{align*}
	e(x) = \Pr(s=1|x,y=1).
	\end{align*}
\end{definition}

When the labeling mechanism is understood, incorporating it during the learning phase enables learning an unbiased classifier from SAR PU data. However, when it is not known, additional assumptions are needed to enable learning~\cite{bekker2019ecml}.

\subsubsection{Probabilistic Gap}
\label{sec:pgpu}
Here, it is assumed that positive examples which resemble negative examples more, are less likely to be labeled. The difficulty of labeling is defined by the \emph{probabilistic gap} $\Delta\Pr(x) = \Pr(y=1|x)-\Pr(y=0|x)$~\cite{he2018idpu}. The labeling mechanism depends on the attribute values $x$ and is therefore a specific case of SAR, which is illustrated in Figure~\ref{fig:sar1}. \revisions{This assumption is satisfied naturally in many applications. Diseases with fewer symptoms are more difficult to diagnose, and users are more likely to click on ads that they are more interested in.}

\begin{definition}[Probabilistic Gap  PU (PGPU)] Labeled examples are a biased sample from the positive distribution, where examples with a smaller probabilistic gap $\Delta\Pr(x)$ are less likely to be labeled. The propensity score is a non-negative, monotone \revisions{increasing} function $f$ of the probabilistic gap $\Delta\Pr(x)$:
	\begin{align*}
		e(x) = f\left(\Delta\Pr(x)\right)=f\left(\Pr(y=1|x)-\Pr(y=0|x)\right)&,&\frac{d}{dt}f(t)\revisions{>}0.
	\end{align*}
\end{definition}

\revisions{
The observed probabilistic gap $\Delta\tilde\Pr(x)=\Pr(s=1|x)-\Pr(s=0|x)$ is related to the real probabilistic gap as follows:
\[
\Delta\tilde\Pr(x) = e(x)(\Delta\Pr(x)+1)-1.
\]
There are two important properties of this relationship.
\begin{enumerate}
	
\item The observed probabilistic gap is always smaller than or equal to the real probabilistic gap:
\[
	\Delta\tilde\Pr(x) \leq \Delta\Pr(x).
\]

\proof{
	\begin{align*}
		\Delta\tilde\Pr(x) 	&=e(x)(\Delta\Pr(x)+1)-1 \\
							&\leq (\Delta\Pr(x)+1)-1 \qquad \#~e(x)\in [0,1]~\textit{and}~\Delta\Pr(x)\geq-1 \\
							&= \Delta\Pr(x).
	\end{align*}
}

From this property it follows that an observed positive probabilistic gap implies a real positive probabilistic gap. This can be used to extract reliable positive examples by selecting examples with an observed positive probabilistic gap~\cite{he2018idpu}.\\

\end{enumerate}
\begin{enumerate}
\setcounter{enumi}{1}
\item Given the probabilistic gap assumption, the observed probabilistic gap maintains the same ordering as the probabilistic gap:
\begin{align}
    \Delta\tilde\Pr(x_1)=\Delta\tilde\Pr(x_2) &\iff \Delta\Pr(x_1)=\Delta\Pr(x_2), \label{eq:pgpu_eq}\\
	\Delta\tilde\Pr(x_1)>\Delta\tilde\Pr(x_2) &\iff \Delta\Pr(x_1)>\Delta\Pr(x_2).\label{eq:pgpu_ineq}
\end{align}

\proof{
The equality of ~\ref{eq:pgpu_eq} is proven by the insight that if two instances have the same probabilistic gaps (i.e., $\Delta\Pr(x_1)=\Delta\Pr(x_2)$), then they must have the same propensity scores, because these are a function of the probabilistic gap $e(x)=f(\Delta(x))$.
\begin{align*}
\Delta\tilde\Pr(x_1) & = f(\Delta\Pr(x_1))(\Delta\Pr(x_1)+1)-1 \\  
                   & = f(\Delta\Pr(x_2))(\Delta\Pr(x_2)+1)-1 \\    
                   & = \Delta\tilde\Pr(x_2).
\end{align*}
The inequality of Equation~\ref{eq:pgpu_ineq} is proven by the insight that under the probabilistic gap assumption, an instance with a larger probabilistic gap $\Delta\Pr(x_1)>\Delta\Pr(x_2)$ has a larger propensity score $e(x_1)=f(\Delta\Pr(x_1))>f(\Delta\Pr(x_2))=e(x_2)$ because the propensity score is a monotone increasing function of the probabilistic gap:
\begin{align*}
\Delta\tilde\Pr(x_1) & = f(\Delta\Pr(x_1))(\Delta\Pr(x_1)+1)-1\\
                     & > f(\Delta\Pr(x_2))(\Delta\Pr(x_2)+1)-1\\
                     & = \Delta\tilde\Pr(x_2).
\end{align*}
}

This property can be used to extract reliable negative examples, by selecting unlabeled examples with an observed probabilistic gap that is smaller than the smallest observed probabilistic gap of the labeled examples~\cite{he2018idpu}.
\end{enumerate}

}

\subsection{Data Assumptions}
The common assumptions about the data distribution are that all unlabeled examples are negative, the classes are separable and the classes have a smooth distribution. 

\subsubsection{Negativity}
The most simple, and most naive, assumption is to assume that the unlabeled examples all belong to the negative class. Despite the fact that this assumption obviously does not hold, it is often used in practice. In the context of knowledge bases, this assumption is commonly referred to as the \emph{closed-world assumption}.
The reason why this assumption is popular is because it enables the use of standard machine learning methods for supervised binary classification~\cite{neelakantan2015compositional}. 
This assumption is simply cited for completeness, and is ignored for the remainder of this survey.

\subsubsection{Separability}
\label{sec:separability}

\begin{figure}[t]
	\centering
	\includegraphics[scale=0.4]{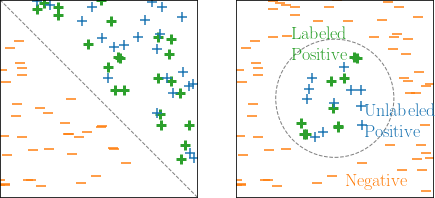}
	\caption[Examples of separable classes]{\textbf{Examples of separable classes.} The first example is linearly separable by a function $f(x_0,x_1) = x_0+x_1$. The second example is separable by a circle, i.e., by a function $f(x_0,x_1) = -\sqrt{x_0^2+x_1^2}$.}
	\label{fig:separability}
\end{figure}

Under the separability assumption, it is assumed that the two classes of interest are naturally separated. This means that a classifier exists that can perfectly distinguish positive from negative examples.  Figure~\ref{fig:separability} shows some examples of separable classes.

\begin{definition}[Separability]
	There exists a function $f$ in the considered hypothesis space that maps all the positive examples to a value that is higher or equal to a threshold $\tau$ and all negative examples to a value that is lower than threshold $\tau$:
	\begin{align*}
	f(x_i) \geq \tau& ,\quad y_i=1\\
	f(x_i) <\tau &,\quad y_i=0.
	\end{align*}
\end{definition}

Under this assumption, the optimal classifier can be found by looking for the classifier that classifies all labeled examples as positive and as few as possible examples as negative~\cite{liu2002partially,blanchard2010semi}. This idea is exploited by the two-step techniques (Section~\ref{sec:twostep}).

\subsubsection{Smoothness}
\label{sec:smoothness}
According to the smoothness assumption, examples that are close to each other are more likely to have the same label.

\begin{definition}[Smoothness]
	If two instances $x_1$ and $x_2$ are similar, then the probabilities $\Pr(y=1|x_1)$ and $\Pr(y=1|x_2)$ will also be similar.
\end{definition}

This assumption allows identifying reliable negative examples as those that are far from all the labeled examples. This can be done by using different similarity (or distance) measures such as tf-idf for text~\cite{li2003learning} or DILCA for categorical attributes~\cite{Ienco2016PUL}. This assumption is important for two-step techniques (Section~\ref{sec:twostep}). It is also used for graph-based approaches~\cite{pelckmans2009transductively,Yu2007PEPUCAG}, local learning~\cite{Ke2017GlobalAL} and to cluster the data into super-instances where all the instances are assumed to have the same label~\cite{li2009positive}.

\subsection{Assumptions for an Identifiable Class Prior}
\label{sec:assumptions_alpha}
The class prior $\alpha=\Pr(y=1)$ can be an important tool for PU learning under the SCAR assumption. Therefore, it would be useful if it could be estimated directly from PU data. Unfortunately, this is an ill-defined problem because it is not identifiable: the absence of a label can be explained by either a small prior probability for the positive class or a low label frequency~\cite{Scott2015ARO}. In order for the class prior to be identifiable,  additional assumption are necessary. This section gives an overview on possible assumptions, listed from strongest to strictly weaker.

\begin{enumerate}
	\item \emph{Separable Classes/Non-overlapping distributions} Here, the positive and negative distributions are assumed not to overlap~\cite{elkan2008learning,Plessis2014ClassPE,Northcutt2017LearningWC}. The positive examples in the unlabeled data are then all those that are likely to be generated by the same distribution as the labeled examples. When all the unlabeled positive examples are identified, class prior estimation becomes trivial.
	\item \emph{Positive subdomain/anchor set} Instead of requiring no overlap between the distributions, it suffices to require a subset of the instance space defined by partial attribute assignment (called the anchor set), to be purely positive~\revisions{\cite{bekker2018aaai,Liu2016ClassificationWN,Plessis2015ClasspriorEF,Scott2015ARO}}. The ratio of labeled examples in this subdomain is equal to the label frequency, while in other parts of the positive distribution, the ratio can be lower.
	\item \emph{Positive function/separability} This  is a more general version of the positive subdomain assumption, where the subdomain can be defined by any function instead of being limited to partial variable assignments~\cite{Ramaswamy2016MixturePE}. When this assumption was introduced, it was named `separability', which we find confusing and thus recommend the more intuitive name `positive function'.
	\item \emph{Irreducibility} The negative distribution cannot be a mixture that contains the positive distribution~\cite{blanchard2010semi,Jain2016EstimatingTC}. All the previous assumption imply irreducibility.
\end{enumerate}

\section{PU Measures}
\label{sec:evalmetrics}

It is non-obvious how to compute most standard evaluation metrics, such as accuracy, $F_1$ score, mean square error, etc. from positive and unlabeled data. This introduces challenges both in terms of model evaluation and hyperparameter tuning. The first attempts for addressing this issue focused on proposing metrics that could be computed based on the total number of examples and the number of positive examples. More recent work has explored hypothesis testing and situations where it may be possible to compute standard metrics. 

\subsection{Metrics for PU Data}
The most commonly used metric for tuning using PU data is based on the $F_1$ score, which is defined as:
\begin{align*}
F_1(\hat \y) &= \frac{2pr}{p+r},
\end{align*}
with precision $p=\Pr(\y=1|\hat\y=1)$ and recall $r=\Pr(\hat\y=1|\y=1)$. Under the SCAR assumption, the recall can be estimated from PU data: $r=\Pr(\hat \y=1|\s=1)$, however, the precision cannot. The $F_1$ score cannot be estimated directly from the PU data, but something similar can be. Note that the $F_1$ score is high when both precision and recall are high. The following performance criterion has the same property and can be estimated from PU data~\cite{lee2003learning}:
\begin{align}
\frac{pr}{\Pr(\y=1)} &= \frac{pr^2}{r\Pr(\y=1)}\nonumber\\
&= \frac{\Pr(\y=1|\hat\y=1)r^2}{\Pr(\hat \y=1,\y=1)}\nonumber\\
&= \frac{r^2}{\Pr(\hat \y=1)}\label{eq:recall}.
\end{align}

\subsection{Hypothesis Testing}
The G-test is and independence test based on mutual information that can be used for structure learning or feature selection. It turns out that the result of observing independence with the G-test is the same from supervised and PU data. However, the power of the test differs with a constant correction factor $\frac{1-\alpha}{\alpha}\frac{\Pr(s=0)}{1-\Pr(s=0)}$. Because the correction factor is a constant that depends on the amount of labeled data, one can calculate how much more data is required to get the desired power~\cite{Sechidis2014StatisticalHT}. The conditional test of independence, which  was used for learning the PTAN trees, has similar properties~\cite{Calvo2007LBC,Sechidis2015MarkovBD}. For feature selection, one is interested in ranking the features in order of mutual information between the features and the label. Interestingly, this order remains the same when the unlabeled examples are considered as negative~\cite{Sechidis2017SimpleSF}.

\subsection{Computing Standard Evaluation Metrics}

More recently, it has been shown that under certain conditions it is possible to compute (bounds on) traditional metrics used to evaluate learned models~\cite{claesen2015assessing,Jain2017RecoveringTC}. Effectively, making the SCAR assumption leads to two important insights. First, by estimating the label frequency or class prior, it is possible to compute the expected number of positive examples in the unlabeled data.  Second, the rank distributions of the observed positives and the positive examples contained within the unlabeled data should be similar.  Combining these two pieces of information enables reasoning about the total number of  positive examples (i.e., the sum of  the observed positives and the expected number of positives in the unlabeled  data) below (above) a given rank. This is precisely the information needed to construct contingency tables, which can be used to derive standard machine learning metrics such as accuracy, the true positive rate, the false positive rate, and precision. Hence, it is possible in this circumstance to report estimates of these metrics.

        \section{PU Learning Methods}
        \label{sec:pumethods}
        
        This section provides an overview of the methods that address PU learning. Most methods can be divided into the following three categories: Two-step techniques, biased learning and class prior incorporation. The two-step technique consists of two steps: 1) identifying reliable negative examples, and 2) learning based on the labeled positives and reliable negatives. Biased learning considers PU data as fully labeled data with class label noise for the negative class. Class prior incorporation modifies standard learning methods by applying the mathematics from the SCAR assumption directly, using the provided class prior. Additionally, methods for learning from relational PU data are discussed.
        
        \subsection{Two-Step Techniques}
        \label{sec:twostep}
        The two-step technique builds on the assumptions of separability and smoothness. Because of this combination, it is assumed that all the positive examples are similar to the labeled examples and that the negative examples are very different from them. Based on this idea, the two-step technique consists of the following steps~\cite{liu2003building}: 
        \begin{description}
        	\item [\textit{Step 1}] Identify reliable negative examples. Optionally, additional positive examples can also be generated~\cite{Fung2006TextCW}.
        	\item [\textit{Step 2}] Use (semi-)supervised learning techniques with the positively labeled examples, reliable negatives, and, optionally, the remaining unlabeled examples.
        	\item [\textit{Step 3 (when applicable)}] Select the best classifier generated in step 2.
        \end{description}
        
        Several methods exist for each one of the steps, which are discussed in the following paragraphs. Despite the possibility of choosing the method freely per step~\cite{liu2003building}, most papers propose a fixed combination of methods, which are listed in  Table~\ref{tab:2step}.

        \begin{table}
        	\centering
        	\caption[Two-step techniques]{\textbf{Two-step techniques.} Despite the possibility of choosing the method freely per step, the following combinations where proposed in the literature. Variations of methods are indicated with $\ast$.}
        	\label{tab:2step}
        	\begin{tabular}{l|lll}
        		Method	& Step 1	& Step 2	& Step 3\\\hline
        		S-EM~\cite{liu2002partially}	& Spy	& EM NB & $\Delta E$\\
        		Roc-SVM~\cite{li2003learning}	& Rocchio	& Iterative SVM & $FNR>5\%$\\
        		Roc-Clu-SVM~\cite{li2003learning}	& Rocchio$^\ast$	& Iterative SVM & $FNR>5\%$\\
        		PEBL~\cite{Yu2002PEBLPE,Yu2004PEBLWP}		& 1-DNF	& Iterative SVM	& Last\\
        		A-EM~\cite{Li2005LearningFP}	& Augmented Negatives 	& EM NB 	& $\Delta F$\\
        		LGN~\cite{Li2007LearningTI}		& Single Negative	&	BN	& / \\
        		PE\_PUC~\cite{Yu2007PEPUCAG} & PE    & (EM) NB   & Unspecified \\
        		WVC/PSOC\cite{Peng2007SVMBA}	&1-DNF$^\ast$	&	Iterative SVM	& Vote \\
        		CR-SVM ~\cite{Li2010NegativeTD}		& Rocchio$^\ast$ &SVM	&/ \\
        		MCLS~\cite{Chaudhari2012LPU}	& k-means	& Iterative LS-SVM	& Last\\
        		C-CRNE~\cite{Liu2014ClusteringbasedMF}	& C-CRNE	& TFIPNDF	&/\\
        		Pulce~\cite{Ienco2016PUL}	& DILCA & DILCA-KNN	&/\\		
        		PGPU~\cite{he2018idpu} & PGPU & biased SVM &/
        	\end{tabular}
        \end{table}

        \paragraph{Step 1: Identifying Reliable Negatives (and Positives)}
        In the first step, unlabeled examples that are very different from the positive examples are selected as reliable negatives. Many methods have been proposed to address this problem. They differ from each other in the way distance is defined and when something is considered as different enough. Many two-step papers addressed text classification problems, therefore, many distance measures originate from that domain \cite{liu2002partially,li2003learning,Yu2004PEBLWP,Li2005LearningFP,Fung2006TextCW,Li2007LearningTI,Li2010NegativeTD,Lu2010SemisupervisedTC,Liu2014ClusteringbasedMF}. The following methods have been proposed to identify reliable negative and possibly positive examples:

        \begin{description}
        	\item [\textit{Spy}] Some of the labeled examples are turned into spies by adding them to the unlabeled dataset. Then, a Naive Bayes classifier is trained, considering the unlabeled examples as negative, and updated once using expectation maximization. The reliable negative examples are all the unlabeled negative examples for which the posterior probability is lower than the posterior probability of any of the spies~\cite{liu2002partially}. For this method, it is important to have enough labeled examples, otherwise the set of spies is too small and hence unreliable.
        	\item [\textit{1-DNF}] First, strong positive features are learned by searching for features that occur more often in the positive data than in the unlabeled data. The reliable negative examples are the examples that do not have any strong positive features~\cite{Yu2004PEBLWP}. Because the requirements for positive features are so weak, there might be too many, resulting in very few reliable negative examples. To resolve this, 1-DNFII proposes to discard positive features with an absolute frequency above some threshold~\cite{Peng2007SVMBA}.
        	\item [\textit{Rocchio}] Based on Rocchio classification, this methods builds a prototype for both the labeled and the unlabeled examples. The prototype is the weighted difference of the mean vector of the tf-idf feature vectors of the objective class and the mean vector of the tf-idf feature vectors of the other class. The unlabeled examples that are closer to the unlabeled prototype than the positive prototype are chosen to be the reliable negatives~\cite{li2003learning}. In addition to Rocchio, k-means clustering can be applied to be more selective: every reliable negative that is closer to a positive prototype than a negative one is removed in this step~\cite{li2003learning}. Another modification with the aim of being more selective only uses potential unlabeled examples, selected using the cosine similarity, for the negative prototype~\cite{Li2010NegativeTD}. Yet another modification is to combine Rocchio with k-means to extract also reliable positive examples in addition to more reliable negatives~\cite{Lu2010SemisupervisedTC}.
        	\item [\textit{PNLH}] The Positive examples and Negative examples Labeling Heuristic\linebreak (PNLH) aims to extract both reliable negative and positive examples. First, reliable negatives are extracted using features that more frequently occur in positive data. Subsequently, the sets of reliable positives and negatives are iteratively enlarged by clustering the reliable negatives. Examples that are close to the positive cluster and to no negative cluster are added to the reliable positives. Examples that are close to a negative cluster and not to the positive one are added to the reliable negatives~\cite{Fung2006TextCW}.
        	\item [\textit{PE}] Positive Enlargement aims to extract reliable negative and positive examples. A graph-based semi-supervised learning method is used to extract reliable positives and Naive Bayes for reliable negatives~\cite{zhou2004learning}.
        	\item [\textit{PGPU}] Under the probabilistic gap assumption (see Section~\ref{sec:pgpu}), all examples with a \revisions{positive} observed probabilistic gap can confidently be considered as \revisions{positive}, and all examples with an observed probabilistic gap that is \revisions{smaller} than the probabilistic gap of any observed positive example can confidently be considered as \revisions{negative}~\cite{he2018idpu}.
        	\item [\textit{k-means}] All the examples are clustered using k-means. Reliable negative examples are selected from the negative clusters as the furthest ones from the positive examples~\cite{Chaudhari2012LPU}.
        	\item [\textit{kNN}] The unlabeled examples are ranked according to their distance to the $k$ nearest positive examples. The unlabeled examples at the greatest distance are selected as reliable negatives~\cite{zhang2009reliable}.
        	\item [\textit{C-CRNE}] Clustering-based method for Collecting Reliable Negative Examples (C-CRNE) is a method that clusters all the examples and takes the clusters without any positive examples as the reliable negatives~\cite{Liu2014ClusteringbasedMF}.
        	\item [\textit{DILCA}] Reliable negatives are selected based on a trainable distance measure DIstance Learning for Categorical Attributes (DILCA), which is designed specifically for categorical attributes~\cite{Ienco2012FromCT}. This distance measure is learned from the positive examples and then used to detect reliable negatives as the furthest examples.
        	\item [\textit{GPU}] Generative Positive-Unlabeled (GPU) learns a generative model for the positive distribution, based on the labeled set of positives. The reliable negatives are the unlabeled examples with the lowest probability of being generated by the generative model. The number of reliable negatives is set to be equal to the number of labeled positives \cite{basile2017de}.
        	\item [\textit{Augmented Negatives}] Instead of selecting reliable negative examples, the unlabeled set is enriched with new examples that are most likely negative. All the unlabeled and added examples are then initialized as negative~\cite{Li2005LearningFP}. This method is intended for the one-class classification setting where the distribution of negative examples can be different at test time.
        	\item [\textit{Single Negative}] This method generates a single artificial negative example. This method is intended for an outlier detection setting where very few negative examples are expected in the unlabeled data~\cite{Li2007LearningTI}.
        \end{description}

        \paragraph{Step 2: (Semi-)Supervised Learning}
        In the second step, the labeled positive examples and reliable negatives are used to train a classifier. Any supervised method, like support vector machines (SVM) or Naive Bayes (NB), can be used for this. Semi-supervised methods, like Expectation Maximization on top of Naive Bayes (EM NB), can also incorporate the remaining unlabeled examples. If semi-supervised methods are used, some methods use the extracted reliable examples from the first step as an initialization that can be changed during the learning process~\cite{liu2002partially,Li2005LearningFP,Chaudhari2012LPU}, while others fix them and only consider the remaining unlabeled examples for possibly belonging to both classes~\cite{li2003learning,Yu2004PEBLWP}. Apart from existing methods, a few custom methods for PU learning have been proposed:
        
        \begin{description}
        	\item [\textit{Iterative SVM}] In each iteration, an SVM classifier is trained using the positive examples and the reliable negatives. The unlabeled examples that are classified as negative by this classifier are then added to the set of reliable negatives for the next iteration~\cite{yu2005single}. 	
        	\item [\textit{Iterative LS-SVM}] In each iteration, a non-linear least Squares SVM (LS-SVM)~\cite{Suykens1999LeastSS} classifier is trained. During the first iteration, the positive and negative examples come from the initialization. In the later iterations, they come from the classification of the previous iteration. In every iteration, the bias is determined by the desired class ratio~\cite{Chaudhari2012LPU}.
        	\item [\textit{DILCA-KNN}] For both the positive and reliable negative examples, a DILCA distance measure is trained~\cite{Ienco2012FromCT}. For each example, the k nearest positives and k nearest reliable negatives are selected and the average distance to those are calculated with the appropriate distance measure. The class is the one for which it has the lowest average distance~\cite{Ienco2016PUL}.
        	\item [\textit{TFIPNDF}] Term Frequency Inverse Positive-Negative Document Frequency is a tf-idf-improved method that weights the terms in documents according to their appearance in positive and negative documents~\cite{Liu2014ClusteringbasedMF}.
        \end{description}
        
        \paragraph{Step 3 (Optional): Classifier selection}
        Expectation Maximization (EM) generates a new model during every iteration. The local maximum to which EM converges might not be the best model in the sequence. Therefore, different techniques have been proposed to select a model from the sequence:
        
        \begin{description}
        	\item [$\mathit{\Delta E}$] The chosen model is the one from the last iteration where the estimated change in the probability of error $\Delta E=\Pr(\hat y_i \neq y)-\Pr(\hat y_{i-1}\neq y)$ is negative, i.e., the last iteration where the model improved~\cite{liu2002partially}.
        	\item [$\mathit{\Delta F}$] The chosen model is the one from the last iteration where the estimated change in the $F_1$ score $\Delta F=F_i/F_{i-1}$ is larger than 1, i.e., the last iteration where the model improved~\cite{Li2005LearningFP}.
        	\item [$\mathit{FNR>5\%}$] Stops iterating if more than $5\%$ of the labeled positive examples are classified as negative~\cite{li2003learning}.
        	\item [\textit{Vote}] All the intermediate classifiers are used and their results are combined through weighted voting. The optimal weights can be found through Particle Swarm Optimization (PSO)~\cite{Peng2007SVMBA}.
        	\item [\textit{Last}] The selected model is the one from the last iteration, when the model has converged or the maximum number of iterations was reached.
        \end{description}
        \vspace{20pt}

        \subsection{Biased Learning}
        \label{sec:penlearning}
        Biased PU learning methods treat the unlabeled examples as negatives examples with class label noise, therefore, this section refers to unlabeled examples as negative. Because the noise for negative examples is a constant, this setting makes the SCAR assumption. The noise  is taken into account by, for example, placing higher penalties on misclassified positive examples or tuning hyperparameters based on an evaluation metric that is suitable for PU data. Usually the misclassification penalties or other hyperparameters are chosen through tuning using Equation~\ref{eq:recall}~\cite{liu2003building,Claesen2015ARE,Zhang2014NonparallelHS,Sellamanickam2011APR} or another measure~\cite{shao2015laplacian}. Alternatively, they are set based on the true class prior~\cite{hsieh2015pu} or so that a balanced classifier is preferred~\cite{mordelet2014bagging,lee2003learning}. This approach has been applied to classification, clustering and matrix completion.
        
        \subsubsection{Classification}
        A large fraction of the biased learning methods are based on \emph{support vector machine (SVM)} methods. The original one is biased SVM which is a standard SVM method that penalizes misclassified positive and negative examples differently \cite{liu2003building}. As an extension, multiple iterations of biased SVM can be executed where misclassified confident unlabeled examples receive an extra penalty~\cite{ke2012building}. Weighted unlabeled samples SVM (WUS-SVM) assigns a weight to each unlabeled example, on top of the class penalty, that indicates how likely this examples is to be negative. The weight is the minimum distance to a positive example~\cite{liu2005partially}.
        
        The noisiness of the negative data makes the learning harder: too much importance might be given to a negative example that is actually positive~\cite{scott2009novelty}. This problem has been addressed by using bagging techniques or using least-square SVMs (LS-SVM)~\cite{Suykens1999LeastSS}. Bagging SVM learns multiple biased SVM classifiers which are trained on the positive examples and a subset of the negative examples~\cite{mordelet2014bagging}. Robust Ensemble SVM (RESVM) builds on bagging SVMs by also resampling the positive examples and using a bootstrap approach~\cite{Claesen2015ARE}. Biased least squares SVM (BLSSVM) is a biased version of LS-SVM, which, additionally, enables local learning by using an extra regularization term that favors close-by examples having the same label, using the smoothness assumption~\cite{Ke2017GlobalAL}. BLSSVM has been extended to MD-BLSSVM by using the Mahalanobis~\cite{Mahalanobis193649} distance instead of the Euclidean distance~\cite{KE2018422}.

        RankSVM (RSVM) is an SVM method that minimizes a regularized \linebreak margin-based pairwise loss~\cite{Sellamanickam2011APR}. In this method, the two classes do not get a different penalty, but the regularization parameter and threshold for classification are set by tuning on Equation~\ref{eq:recall}.
        Other hyperplane optimization methods are Biased Twin SVMs~\cite{Xu2014LearningWP}, nonparallel support vector vector machines (NPSVM)~\cite{Zhang2014NonparallelHS}, and the Laplacian Unit-Hyperplane classifier (LUHC)~\cite{shao2015laplacian}.
        
        Weighted \emph{logistic regression} favors correct positive classification over correct negative classification by giving larger weights to positive examples~\cite{lee2003learning}. The positive examples are weighted by the negative class prior $\Pr(s=0)$ and the negative examples by the positive class prior $\Pr(s=1)$. They show that as a result, the conditional probability that a positive example belongs to the positive class is larger than 0.5 while a negative example will have a conditional probability smaller than 0.5. In principle, a correct classifier would thus be learned. However, when the classes are not separable, the overlapping parts of the instance space might be attributed to the wrong class. This is because the weighting is equivalent to setting the target probability threshold for the non-traditional classifier to $c\Pr(y=1)$, while it should be $0.5c$~\cite{Elkan2001TheFO}. Separable classes can handle this by having ${0,1}$ probabilities, but non-separable classes are only correctly classified if they are balanced. This is discussed in more detail in Section~\revisions{\ref{sec:preprocessing}}.
        
        \subsubsection{Clustering}
        Topic-Sensitive pLSA (probabilistic latent semantic analysis) is a weighted \emph{constraint clustering} method that introduces must-link constraints between pairs of positive examples and cannot-link constraints between examples from different classes~\cite{Zhou2010LearningWP}. The must-link constraints have stronger weights than the cannot-link constraints. This method is expected to work well when the number of labeled positive examples is small.
        
        \subsubsection{Matrix Completion}
        Binary \emph{matrix completion} can also be seen as a PU learning problem: the ones in the matrix are the known positives and the zeros are unlabeled~\cite{hsieh2015pu}. They assume that in reality, there is a probability matrix of the same size which generated the complete binary matrix. Two binary matrix generation settings are considered: 1) The non-deterministic setting where the complete binary matrix was generated by sampling from the probability matrix, and  2) The deterministic setting where the complete binary matrix was generated by thresholding the probability matrix. The observed matrix is generated by uniform sampling from the complete binary matrix.
        
        In the non-deterministic setting, it is possible to recover the probability matrix, if the true class prior is known. To this end, Shifted Matrix Completion (ShiftMC) minimizes an unbiased estimator for the mean square error loss. This is a special case of the general empirical-risk-minimization based method for incorporating the class prior by preprocessing the data (see Section~\ref{sec:preprocessing}).
        
        In the deterministic setting, the probability matrix cannot be recovered, but the complete binary matrix can. To this end, the matrix factorization method Biased Matrix Completion (BiasMC) penalizes misclassified positives more than misclassified negatives. The penalties are derived from the class prior. Section~\ref{sec:preprocessing} shows how this is a special case of the rebalancing method for incorporating the class prior by preprocessing the data. An extension to BiasMC for graphs uses the additional information that neighbors are likely similar~\cite{natarajan2015pu}.

        \subsection{Incorporation of the Class Prior}
        \label{sec:classpriorincorp}
        
        Under the SCAR assumption, the class prior can be used. There are three categories of methods: postprocessing, preprocessing and method modification. Postprocessing trains a non-traditional probabilistic classifier by considering the unlabeled data as negative and modifies the output probabilities, preprocessing changes the dataset by using the class prior, and method modification modifies the methods to incorporate the class prior.
        
        Remember from Section~\ref{sec:relalphac} that knowing the class prior is equivalent to knowing the label frequency $c$, which is the proportion of labeled positive examples $c=\Pr(s=1)/\alpha$. The class prior can be determined using methods discussed in Section~\ref{sec:back_alphaest} or it can be tuned using evaluation metrics for PU data, which are discussed in Section~\ref{sec:evalmetrics}.
        
        Under the SAR assumption, in a similar fashion, the propensity score can be incorporated to enable learning. Currently, this has only been explored for the empirical-risk-minimization-based preprocessing method.
        
        \subsubsection{Postprocessing} 
        \label{sec:postprocessing}
        The probability of an example being labeled is directly proportional to the probability of that example being positive, with the label frequency $c$ as the proportionality constant:
        \begin{align*}
        \Pr(s=1|x)=c\Pr(y=1|x).
        \end{align*}
        From this result, it follows directly that a non-traditional probabilistic classifier that is trained to predict $\Pr(s=1|x)$ by considering the unlabeled data as negative can be used to predict the class probabilities $\Pr(y=1|x)=\frac{1}{c}\Pr(s=1|x)$~\cite{elkan2008learning}. Alternatively, when the probabilities are of no importance, the non-traditional classifier can be used directly by changing the target probability threshold $\tau$ to $\tau^{PU}=c\tau$. The commonly used $\tau=0.5$ then results in the decision function $\Pr(s=1)>0.5c$. This is equivalent to the decision function  $\text{sgn}(\Pr(y=1|x)-\Pr(y=0|x))=\text{sgn}(\frac{2-c}{c}\Pr(s=1|x)-\Pr(s=0|x))$ from Zhang and Lee~\cite{zhang2005simple}.

        \subsubsection{Preprocessing}
        \label{sec:preprocessing}
        The goal of preprocessing, is to create a new dataset from a PU dataset, which can be used by methods that expect fully supervised data to train the best possible model for the PU data. The proposed methods can be ordered into three categories: rebalancing methods, methods that incorporate the label probabilities and, empirical-risk-minimization-based methods.
        
        \paragraph{Rebalancing Methods}
        \label{sec:rebalancing}
        As seen before, a non-traditional classifier, trained on the positive and unlabeled data, gives the same classification as a traditional classifier, if the target probability threshold $\tau$  is set appropriately. Instead of changing the threshold, the rebalancing method from Elkan~\cite{Elkan2001TheFO} can be employed to weight the data so that the classifier trained on the weighted data will give the same classification with the same target probability threshold as the traditional classifier. Given the target probability threshold for the traditional classifier $\tau$, the target probability threshold for the non-traditional classifier would be $\tau^{PU}=c\tau$. To move the target probability from $\tau$ to $\tau^{PU}$ in the non-traditional classifier, the data needs to be weighted as follows:
        \begin{align*}
        w^+ &=\tau(1-\tau^{PU}) 	& w^- &= (1-\tau)\tau^{PU}\\
        &=\tau(1-c\tau)&&=(1-\tau)c\tau\\
        &=(1-c\tau)&&=(1-\tau)c,
        \end{align*}
        where $w^+$ and $w^-$ are the weights for positive and negative examples respectively. In the last step, both weights were divided by $\tau$ to simplify the formula as this does not affect the learning result. When the target probability is $\tau=0.5$, this reduces to 
        \begin{align*}
        w^+ &= 1-c/2	& w^- &= c/2,
        \end{align*}
        which is equivalent to the result used for BiasMC~\cite{hsieh2015pu}. If the true class prior is $\alpha=0.5$, the result reduces to
        \begin{align*}
        w^+ &= 1-c\alpha	& w^- &= c\alpha\\
        w^+ &= Pr(s=0)	& w^- &= \Pr(s=1)
        \end{align*}
        which are the weights used for weighted logistic regression~\cite{lee2003learning}.
        
        Rank Pruning was proposed to be more robust to noise. To this end, it first cleans the data based on the class prior and the expected positive label noise (both of which are estimated in a first phase, see Section~\ref{sec:back_alphaest}), with the goal of only keeping confident positive and negative examples. The confident examples are then weighted to get the correct class prior~\cite{Northcutt2017LearningWC}.
        
        Rebalancing methods are only appropriate when one is interested in classification on the given target threshold $\tau$, but not for returning the unbiased estimates of the probability $\Pr(y=1|x)$.
        
        \paragraph{Incorporation of the Label Probabilities}
        
        Elkan and Noto~\cite{elkan2008learning} proposed to duplicate the unlabeled examples to let them count partially as positive and partially as negative. The weights are the probabilities of the unlabeled examples being positive and negative respectively. The labeled examples are certain to be positive and are therefore added as positive examples with weight 1. The probability for an unlabeled example to be positive is
        \begin{align*}
        \Pr(y=1|s=0,x) = \frac{1-c}{c}\frac{\Pr(s=1|x)}{1-\Pr(s=1|x)}.
        \end{align*}
        To generate the weighted dataset like this, first a non-traditional classifier to predict $\Pr(s=1|x)$ needs to be trained.
        
        \paragraph{Empirical-Risk-Minimization Based Methods} 
        \label{sec:erm}
        
        The goal of preprocessing the PU data is that the classifier learned from the resulting dataset is expected to be equal to the classifier trained from a fully labeled dataset. In an empirical risk minimization framework, this means finding the classifier $g$ that minimizes the risk, given some loss function $L$
        \begin{align*}
            R(g) &= \alpha \mathbb{E}_{f_+}\left[L^+(g(x))\right]+(1-\alpha)\mathbb{E}_{f_-}\left[L^-(g(x))\right],
        \end{align*}
        where $L^+(\hat y)$ and $L^-(\hat y)$ are the losses for positive and negative examples respectively. The following are some popular loss functions:
        \begin{align*}
        \text{MAE}: &\qquad L^+(\hat y) = 1-\hat y && L^-(\hat y) = \hat y,\\
        \text{MSE}: &\qquad L^+(\hat y) = (1-\hat y)^2 && L^-(\hat y) = \hat y ^2 \\
        \text{Log Loss}: &\qquad L^+(\hat y) = -\ln \hat y && L^-(\hat y) = -\ln(1-\hat y).
        \end{align*}
        Empirical-Risk-Minimization based-methods, such as SVMs, logistic regression and deep networks, minimize the empirical risk, which is calculated from the data as follows:
        \begin{align}
            \hat R (g|\x,\y) &= \alpha \frac{1}{|\y=\mathbf{1}|}\sum_{x:\x|\y=\mathbf{1}}L^+(g(x)) + (1-\alpha) \frac{1}{|\y=\mathbf{0}|}\sum_{x:\x|\y=\mathbf{0}}L^-(g(x))\nonumber\\
            &= \frac{1}{|\y|}\left( \sum_{x:\x|\y=\mathbf{1}}L^+(g(x)) + \sum_{x:\x|\y=\mathbf{0}}L^-(g(x)) \right). \label{eq:er}
        \end{align}
        
        In PU data, the empirical risk cannot be calculated directly because not all the class values are observed. However, the PU data and the labeling mechanism can be used to create a new, weighted dataset that is expected to give the same empirical risk as the fully labeled data. Next, the risk is rewritten in terms of expectations over the labeled and unlabeled distributions. Then, it is shown how to create the data which gives the same empirical risk when using the standard formula~\ref{eq:er} which is used by standard methods and implementations.
        
        The expectation over the negative distribution can be formulated in terms of expectations over the general and the positive distributions, using Equation~\ref{eq:distr}. The expectation over the positive distribution can be formulated in terms of an expectation over the labeled distribution and the propensity score, using Equation~\ref{eq:labdistr}:
        
        \begin{align*}
            R(g) &= \alpha \mathbb{E}_{f_+}\left[L^+(g(x))\right]+(1-\alpha)\mathbb{E}_{f_-}\left[L^-(g(x))\right] \\
             &= \alpha \mathbb{E}_{f_+}\left[L^+(g(x))\right]+\mathbb{E}_{f}\left[L^-(g(x))\right] - \alpha\mathbb{E}_{f_+}\left[L^-(g(x))\right] \\
             &= \alpha \mathbb{E}_{f_+}\left[L^+(g(x))-L^-(g(x))\right]+\mathbb{E}_{f}\left[L^-(g(x))\right] \\
             &= \alpha \mathbb{E}_{f_l}\left[\frac{c}{e(x)}\left(L^+(g(x))-L^-(g(x))\right)\right]+\mathbb{E}_{f}\left[L^-(g(x))\right].
        \end{align*}
        
        In the case-control scenario, the expectation over the general distribution can simply be replaced by the expectation over the unlabeled distribution. Therefore, the empirical risk is calculated as follows:
        \begin{align*}
            \hat R(g|\x,\s) &= \frac{\alpha}{|\s=\mathbf{1}|}\sum_{x:\x|\s=\mathbf{1}}\left(\frac{c}{e(x)}\left(L^+(g(x))-L^-(g(x))\right) \right) \\&\quad+ \frac{1}{|\s=\mathbf{0}|}\sum_{x:\x|\s=\mathbf{0}}L^-(g(x)).&&\textit{\# case-control}
        \end{align*}
        Hence, the new dataset is created by adding all unlabeled examples as negative with weight $\frac{1}{|\s=\mathbf{0}|}$, and all labeled examples both as positive with weight $\frac{1}{|\s=\mathbf{1}|}\frac{\alpha c}{e(x)}$ and as negative with weight $-\frac{1}{|\s=\mathbf{1}|}\frac{\alpha c}{e(x)}$.
        
        For the single-training-test scenario, the general distribution is a combination of the labeled and unlabeled distributions (Equation~\ref{eq:sts_tot}), which reduces the risk to:
        \begin{align*}
            R(g) &= \alpha c \mathbb{E}_{f_l}\left[\frac{1}{e(x)}L^+(g(x))+
            \left(1-\frac{1}{e(x)}\right)L^-(g(x))\right]\\
            &\quad+(1-\alpha c)\mathbb{E}_{f_u}\left[L^-(g(x))\right].\qquad\qquad\textit{\# single-training-set}
        \end{align*}
        And the empirical risk to:
        \begin{align*}
            \hat R(g|\x,\s) &= \frac{\alpha c}{|\s=\mathbf{1}|} \sum_{x:\x|\s=\mathbf{1}}\left(\frac{1}{e(x)}L^+(g(x))+    \left(1-\frac{1}{e(x)}\right)L^-(g(x))\right)\\
            &\quad+\frac{1-\alpha c}{|\s=\mathbf{0}|}\sum_{x:\x|\s=\mathbf{0}}\left(L^-(g(x))\right)\\
            &= \frac{1}{|\s|}\Bigg( \sum_{x:\x|\s=\mathbf{1}}\left(\frac{1}{e(x)}L^+(g(x))+    \left(1-\frac{1}{e(x)}\right)L^-(g(x))\right)\\ &\qquad\qquad+\sum_{x:\x|\s=\mathbf{0}}\left(L^-(g(x))\right)\Bigg).\qquad\qquad\textit{\# single-training-set}
        \end{align*}
        Hence, the new dataset is created by adding all unlabeled examples as negative with weight 1 and all labeled examples both as positive with weight $\frac{1}{e(x)}$ and as negative with weight $(1-\frac{1}{e(x)})$.
        
        This general weighting method was proposed in the single-training-set scenario as the first SAR PU learning method~\cite{bekker2019ecml} but it already existed before under the SCAR assumption~\cite{steinberg1992estimating,Plessis2015ConvexFF,Kiryo2017PositiveUnlabeledLW}. The ShiftMC method for matrix completion is also a special case of this method under the SCAR assumption, using the MSE loss~\cite{hsieh2015pu}.
        
        du Plessis et al. (2014) proposed another risk estimator, which simply reweights the examples and does not introduce duplicates~\cite{Plessis2014AnalysisOL}. However,the derivation is limited to 0-1 predictions and the method is biased, unless the loss functions sum to one $L^+(\hat y)+L^-(\hat y)=1$, which can only be achieved with non-convex functions. 
        
        \subsubsection{Method Modification} Many machine learning methods are based on counts of positive and negative examples in subsets of the data. The counts are used to calculate (conditional) probabilities, support, coverage or other metrics that are used to make decisions or set parameters. The counts can be estimated using the same rationale as were used for data weighting~\cite{elkan2008learning}.
        
        The PU tree learning algorithm POSC4.5, one of the first PU learning methods, needs the count of positive and negative examples in every considered split for the three. They estimate the number of positives in node $i$ as $\hat P_i=\min\{\frac{1}{c}L_i,T_i\}$ and the negatives as $\hat N_i=T_i-\hat P_i$, where $L_i$ and $T_i$ are the number of labeled and total examples in that node~\cite{denis2005learning}. This corresponds to empirical-risk-minimization-based weighing.
        
        Ward et al.~\cite{ward2009presence} proposed an expectation maximization method on top of logistic regression. The expectation step finds the expected class labels and the maximization step trains the logistic regression model using the expected class labels, followed by rebalancing the model according using the class prior.
        
        For Naive Bayes methods, the probabilities $\Pr(x^{(i)}|y)$, with $x^{(i)}$ the $i$th attribute of $x$, are key. For $y=1$, these can be directly estimated from the labeled data as
        \begin{align}
        \label{eq:condy1}
        \Pr(x^{(i)}|y=1)=\Pr(x^{(i)}|s=1),
        \end{align}
        and for $y=0$ these can be calculated, somewhat less straightforwardly, as follows:
        \begin{align}
        \label{eq:condy0}
        \Pr(x^{(i)}|y=0)&=\frac{\Pr(x^{(i)})-\alpha\Pr(x^{(i)}|y=1)}{1-\alpha} .
        \end{align}
        This insight was used to develop PNB, the first Naive Bayes algorithm for PU learning~\cite{Denis2003TextCA}. It was originally proposed for document classification, but was later generalized to general discrete attributes and incorporate the of Laplace correction~\cite{Calvo2007LBC}. In that same paper an averaging method is presented that can incorporate a distribution over the class prior instead of an exact value. Positive Tree Augmented Naive Bayes (PTAN) builds further on PNB, but also needs to calculate the conditional mutual information between variables $i$ and $k$ for structure learning:
        \begin{align*}
        \sum_j \sum_l \Pr(x^{(i)}=j,x^{(k)}=l,y=1)\log\frac{\Pr(x^{(i)}=j,x^{(k)}=l|y=1)}{\Pr(x^{(i)}=j|y=1)\Pr(x^{(k)}=l|y=1)}&\\
        +\Pr(x^{(i)}=j,x^{(k)}=l,y=0)\log\frac{\Pr(x^{(i)}=j,x^{(k)}=l|y=0)}{\Pr(x^{(i)}=j|y=0)\Pr(x^{(k)}=l|y=0)}&,
        \end{align*}
        all these probabilities can be calculated by using Equations~\ref{eq:condy1}, ~\ref{eq:condy0}, and:
        \begin{align*}
        \Pr(x^{(i)}=j,x^{(k)}=l,y=1) &= \alpha \Pr(x^{(i)}=j,x^{(k)}=l|s=1)\\
        \Pr(x^{(i)}=j,x^{(k)}=l,y=0) &= (1-\alpha)\Pr(x^{(i)}=j,x^{(k)}=l|y=0).
        \end{align*}
        
        Similarly, PU learning methods have been proposed for other Bayesian classifiers. Averaged One-Dependence Estimator (AODE) ~\cite{Webb2005NotSN} has been extended to PAODE, Hidden Naive Bayes (HNB)~\cite{Jiang2009ANB} to PHNB, and Full Bayesian network Classifier (FBC)~\cite{Su2006FullBN} to PFBC~\cite{He2011BCP}. Some of these methods were further extended to uncertain Bayesian methods, where the attribute values are uncertain: UPNB~\cite{He2010NaiveBC} and UPTAN~\cite{Gan2017BBN}, where this last method uses Uncertain Conditional Mutual Information (UCMI) for structure learning~\cite{liang2012learning}.

        \subsection{Relational Approaches}
        A common task for relational data is to complete automatically constructed knowledge bases or networks by finding new relationships. This task can be seen as PU learning, because everything that is already in the knowledge base or network is known to be true and everything that can possibly be added is unlabeled. Most methods make the \emph{closed-world} assumption and learn models by assuming everything that is not in the knowledge base is negative. However, a few methods have been proposed that do make the \emph{open-world} assumption, which makes it explicit that the data is incomplete.
        
        When the SCAR assumption holds in the relational PU data, then, relational versions of classic class prior incorporation methods can be used to enable learning~\cite{bekkerilp17}. TI$c$ER, a relational version of TI$c$E (Section~\ref{sec:tice}) can estimate the class prior directly from the relational PU data.
        
        The PosOnly setting of the relational rule learning system Aleph~\cite{srinivasan2001aleph} makes the separability assumption and looks for the simplest theory that covers all positive examples and introduces as few new facts as possible~\cite{Muggleton:1996}.
        
        RelOCC is a relational one-class classification method which, based on the smoothness assumption, introduces a tree-based distance method~\cite{Khot2014RelationalOC}. They do not use unlabeled examples at training time, so, although related, it is not truly PU learning.
        
        The AMIE+ rule learning system for knowledge base completion introduces the partial completeness assumption. It assumes that if for a subject and relationship at least one object is known, then all objects for this subject and relationship are known. For example, if \texttt{taughtby(bigdata,jesse)}, then it is assumed that the knowledge base contains all Jesse's classes. Using the partial completeness assumption, the confidence of potential rules can be estimated more precisely~\cite{Galarraga2015}. The RC confidence score makes an even more precise estimate, by making a rule-specific SCAR assumption and taking the expected relation cardinalities, i.e., the number of objects/subjects per subject/object and rule combination, into account \cite{zupanc2018estimating}.
        
        PULSE, a relational PU learning algorithm for disjunctive concepts was  proposed in the context of relational grounded language learning~\cite{Blockeel2017PUlearningDC}. In their setting, the positive class can have a limited number of $k$ subclasses. They assume that for each subclass, the SCAR assumption holds, but do not necessary have the same label frequencies.

        \subsection{Other Methods}
        For completeness, this section lists PU methods that do not fit in any of the considered categories.
        \begin{description}
        	\item [\textit{Generative Adversarial Networks (GANs)}] have recently been introduced for PU learning, where they can model the positive and negative distributions~\cite{Hou2018GenerativeAP,Chiaroni2018LearningWA}.
        	\item [\textit{Co-training}] is a semi-supervised learning technique that learns two models, based on two views of the data, where the goal is to find two models that agree~\cite{Blum1998CombiningLA}. This idea has been applied to PU learning as well~\cite{Denis2003TextCA,Zhou2012MultiviewPA}.
        	\item [\textit{Data stream classification}] with PU data has been addressed by multiple \linebreak works~\cite{li2009positive,nguyen2011positive,Qin2012LearningFD,liang2012learning,Chang2016PositiveUnlabeledLI}.
        	\revisions{
        	\item[\textit{Expectation Maximization} (EM)] can be used for SAR PU data with the additional assumption that the propensity scores only depend on a known subset of the attributes. An EM approach is then used to simultaneously train the classifier and a model for estimating the propensity scores~\cite{bekker2019ecml}.
        	}
        \end{description}

\revisions{
\subsection{Comparison of PU Learning Methods}
The primary consideration for choosing a PU learning method is to ascertain which assumptions are mostly likely to hold for the application at hand. 
If separability holds, then this would favor the use of two-step techniques. If SCAR holds, then one would use biased learning or methods that incorporate the class prior. If both separability and SCAR hold, the choice depends on how clearly separated the two classes are. If the classes are separable, but very close to each other, separating the two classes correctly is hard for two-step techniques, so exploiting SCAR is likely more effective. However, if the classes are very clearly separated, the two-step techniques are favored, because, given a clear separation, they are more robust against deviations from the SCAR assumption. Currently, not many methods exist that are tailored towards the SAR and PGPU assumptions. Currently, the only PGPU method is a two-step technique that also assumes separability~\cite{he2018idpu}. Note that this method is preferred to  other two-step techniques, because it builds on the PGPU assumption to find the decision boundary.

If one is interested in unbiased estimates of the true probabilities $\Pr(y=1|x)$ under the SCAR or SAR assumption, then empirical-risk estimation methods should be considered: ERM data reweigthing (Section~\ref{sec:erm}), ShiftMC~\cite{hsieh2015pu}, or POS4.5~\cite{denis2005learning}. The downside of ERM data reweighting is the use of negative weights, which not all classifiers and implementations can handle. The Naive Bayes method PNB~\cite{Denis2003TextCA} and it extensions also output unbiased probabilities. Rebalancing the data (Section~\ref{sec:rebalancing}), or rebalancing/penalizing the classes in biased learning (Section~\ref{sec:penlearning}) are not suited for unbiased probabilities, but are expected to find the correct decision boundary.

Rebalancing and class prior incorporation methods are sensitive to the SCAR assumption. Ensemble methods provide more robustness~\cite{claesen2015robust,mordelet2014bagging}. Alternatively, the smoothness assumption can be leveraged to relax the SCAR assumption~\cite{Ke2017GlobalAL,ke2012building,liu2005partially,Sellamanickam2011APR}.

}

\section{Class Prior Estimation from PU Data}
\label{sec:back_alphaest}
Knowledge of the class prior significantly simplifies PU learning under the SCAR assumption. Therefore, it is very useful to estimate it from PU data directly. To this end, a number of methods have been proposed.

\subsection{Non-traditional Classifier}
 When the classes are separable, in principle a non-traditional classifier $g(x)$ that predicts $\Pr(s=1|x)$ can be trained that maps all negative examples to 0 and all positive examples to $\Pr(s=1|y=1)=c$. Based on this insight, Elkan and Noto~\cite{elkan2008learning} suggest to train a classifier on part of the data while keeping a separate validation set. Then, they estimate the label frequency as the average predicted probability of a labeled validation set example~\cite{elkan2008learning}. \revisions{This method requires} well-calibrated probabilistic classifiers. Methods such as Platt scaling~\cite{platt1999probabilistic}, isotonic regression~\cite{Zadrozny2002TransformingCS} or beta calibration~\cite{Kull2017BetaCA} can be used to calibrate classifiers that do not output well-calibrated probabilities. \revisions{Rank pruning is a more robust method based on a non-traditional classifier $g$ that is based on confident examples: an example $x$ is confidently positive when $g(x)\geq \Pr(\hat{s}=1|s=1)$, with $\hat{s}$ the classification by $g$~\cite{Northcutt2017LearningWC}. The label frequency is calculated from the labeled and unlabeled confident positive examples. This estimation is expected to be correct, as long as the confident positive examples contain no negative examples. Therefore, the method is more robust with regard to the calibration of $g$ and class overlap in the low probability regions. Additionally, rank pruning can handle negative examples that are wrongly labeled in a similar way.}
 
\revisions{
Another method based on a non-traditional classifier uses the insight that the probability $\Pr(s=1|x)=c \Pr(y=1|x)$, which is estimated by $g(x)$, is equal to the label frequency $c$ when the true conditional class probability is $\Pr(y=1|x)=1$~\cite{Liu2016ClassificationWN}. Under the positive subdomain assumption, there will be instances $x$ for which $\Pr(y=1|x)=1$ and hence, the label frequency can be estimated as $c=\max{g(x)}$.
}

\subsection{Partial Matching} 
\begin{figure}[t]
	\centering
	\includegraphics[scale=0.4]{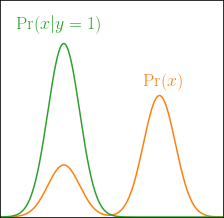}
	\caption[Partial matching]{\textbf{Partial matching.} The goal of partial matching is to find the class prior $\alpha$ that minimizes the divergence between the scaled distributions. This figure is based on Figure 1 in \cite{Plessis2014ClassPE}.}
	\label{fig:partialmatching}
\end{figure}

The partial matching approach assumes non-overlapping classes. It uses a density estimation method to estimate the positive distribution, based on the labeled examples, and the complete distribution, based on all the data~\cite{Plessis2014ClassPE}. The class prior is found by minimizing the difference between the scaled positive distribution, where the scale factor is the class prior. The method is illustrated in Figure~\ref{fig:partialmatching}.

\begin{figure}[t]
	\centering
	\includegraphics[scale=0.4]{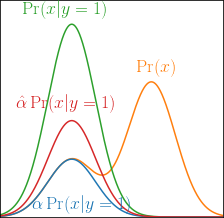}
	\caption[Partial matching with overlap]{\textbf{Partial matching with overlap.} When the classes overlap, the original partial mapping method would result in an overestimate for alpha $\hat\alpha>\alpha$, like the red line. Using a penalized divergence makes sure that the $\alpha$-scaled positive distribution does not surpass the total distribution.}
	\label{fig:partialmatchingoverlap}
\end{figure}

The partial matching approach does not work well when the positive and negative distribution overlap. In this case, the correct class prior would give a large divergence in the regions with overlap. By minimizing the divergence, these regions will favor an overestimate of the class prior. To relax the non-overlapping distributions assumption to the positive subdomain assumption, penalized divergences were introduced~\cite{Plessis2015ClasspriorEF}. These give higher penalties to class priors that result in $\alpha\Pr(x|y=1)>\Pr(x)$  for some $x$. Intuitively, this finds the class prior that scales the positive distribution as closely to the total distribution, without ever surpassing it. The method is illustrated in Figure~\ref{fig:partialmatchingoverlap}

\subsection{Decision Tree Induction}
\label{sec:tice}
Tree Induction for $c$ Estimation (TI$c$E) estimates the label frequency $c$  under the positive subdomain assumption~\cite{bekker2018aaai}. It makes the observation that the label frequency remains the same when considering a subdomain of the data and that the fraction of labeled examples in that subdomain provides a natural lower bound on the label frequency. Using a decision tree induction method, it searches for the subdomain that implies the largest lower bound and returns that as the label frequency estimate. Under the positive subdomain assumption, this lower bound is indeed expected to be the label frequency. \revisions{This method is closely related to the last non-traditional classifier method~\cite{Liu2016ClassificationWN} but differs in that it is more robust and faster. It is more robust because it takes the maximum over sets of instances (subdomains) as opposed to single instances. It is faster because it does not need to train a full tree and instead concentrates on the branches that can give a stricter lower bound.}

\subsection{Receiver Operating Characteristic (ROC) Approaches}
In the ROC setting, one aims to maximize the true positive rate $\text{TPR}=\Pr(\hat y=1|y=1)$ while minimizing the false positive rate $\text{FPR}=\Pr(\hat y=1|y=0)$. The TPR can be calculated in PU data, by using the labeled positive set. While the FPR cannot be calculated from PU data, for a given TPR, minimizing the FPR within a hypothesis space $\mathcal{H}$ is equivalent to minimizing the probability of predicting the positive class $Pr(\hat y=1)$:
\begin{align*}
\min_{\hat y:\mathcal{H},\text{TPR}} \Pr(\hat y=1) &=\min_{\hat y:\mathcal{H},\text{TPR}} \alpha \Pr(\hat y=1|y=1)+(1-\alpha)\Pr(\hat y=1|y=0)\\
&=\min_{\hat y:\mathcal{H},\text{TPR}} \alpha \text{TPR}+(1-\alpha)\Pr(\hat y=1|y=0)\\
&= \alpha \text{TPR}+(1-\alpha)\min_{\hat y:\mathcal{H},\text{TPR}}\Pr(\hat y=1|y=0).\\
\end{align*}
If classifier $f$ exists that minimizes the FPR to zero, then the class prior can be calculated as $\alpha=\Pr(f=1)/TPR=\Pr(f=1)/\Pr(f=1|s=1)$. In fact, for any classifier $f$, this is an upper bound:
\begin{align*}
\alpha \geq \frac{\Pr(f=1)}{\Pr(f=1|s=1)}.
\end{align*}
As a result, maximizing $\Pr(f=1)/\Pr(f=1|s=1)$ over the space of all classifiers gives the class prior~\cite{blanchard2010semi}. This result is valid under the irreducibility assumption. However, without extra assumptions, infinite examples are required for convergence. The stricter positive subdomain assumption allows for practical algorithms. Scott~\cite{Scott2015ARO} implements this idea by building a conditional probability classifier. The same idea is approached from a different angle by Jain et al.~\cite{Jain2016NonparametricSL,Jain2016EstimatingTC}. They use $k$-kernel density estimation to approximate the positive and total distributions, given different values for the class prior $\alpha$, in a second step, they select $\alpha$ as the largest value (i.e., minimal $\Pr(\hat y=1)$ and thus minimal FPR) that results in the optimal log likelihood for both densities (i.e., maximal TPR).

\subsection{Kernel Embeddings}

All previous methods, except TI$c$E, aim to model the entire domain with either discriminative or generative models. However, this might be overkill for estimating one constant, especially since the label frequency is equal for every example. Based on this insight, a class prior estimation method using kernel embeddings is proposed that aims to separate part of the positive distribution from the total distribution, under the positive function assumption. This means that they look for functions that map all negative examples to zero. Given a class prior, the minimal proportion from the negative distribution that is selected by any function is estimated. The class prior is the largest value for which that proportion is below a given threshold~\cite{Ramaswamy2016MixturePE}.

\subsection{Other Sources For the Class Prior} 
Estimating the class prior from PU data is hard. Therefore, it can be useful to obtain it in another way. For some domains, the class prior can be known from domain knowledge or previous studies. If there is access to a smaller dataset for the same domain that does have both possible and unlabeled labels, these can be used to estimate the class prior from. Or finally, one can just not estimate it but treat is as a hyperparameter and use a validation set and tune for it using a PU evaluation metric from Section~\ref{sec:evalmetrics}.

\revisions{

\subsection{Comparison of Prior Estimation Methods}
It is natural to wonder about the relative strengths and weaknesses of the various approaches for estimating the class prior. Whether a particular approach is suitable for a problem will depend on the assumptions underpinning the approach and how well they match the problem at hand. The non-traditional classifier ~\cite{elkan2008learning,Northcutt2017LearningWC} and some partial matching~\cite{Plessis2014ClassPE} approaches make the assumption that the positive and negative example are separable. It is unlikely that this assumption will hold in practice. It is possible to relax this restriction for the partial matching~\cite{Plessis2015ClasspriorEF} approach such that only a positive subdomain is assumed. Moreover, this work is supported by theoretical analysis in terms of uniform deviation bounds and error estimation bounds. The decision tree approach TI$c$E and Jain et al.'s ROC approach~\cite{Jain2016NonparametricSL} also make this same assumption, but do not provide guarantees in terms of convergence to the true estimate.  The kernel embedding approaches KM1 and KM2~\cite{Ramaswamy2016MixturePE} make the even less restrictive positive function assumption. Moreover, the work provides a proof that their algorithm for estimating the prior converges to the true prior under certain assumptions.

Empirically, the comparisons among these approaches tend to focus on idealized conditions on artificially constructed PU data. Hence, which approach is best in practice is still an important open issue. That being said, there are still some insight to be gleaned based on several recent studies. Bekker and Davis~\cite{bekker2018aaai} compared canonical examples of each of aforementioned classes of approaches for estimating the class prior (apart from the techniques in Subsection 6.5). Using a small benchmark (11 datasets)  under a number of different SCAR settings, they found that the kernel embedding approach KM2~\cite{Ramaswamy2016MixturePE} and TI$c$E~\cite{bekker2018aaai} produced the most accurate estimates on SCAR PU data. TI$c$E conferred the advantage of being significantly faster at estimating the class prior. In fact, it was only feasible to run KM2 on small subsets of the data. Of course, KM2 offers the advantage of having stronger theoretical underpinnings. Moreover, recently it was shown that KM2 results in more accurate classifier performance than TI$c$E on SAR PU data~\cite{bekker2019ecml}. 
}

\section{Sources of PU Data and Applications}
\label{sec:sourcesandapplications}
There are many classification situations where PU data naturally occurs and various machine learning tasks can be phrased as PU learning problems. The following subsection lists some of these situations and tasks. Next, applications that were explicitly addressed as PU learning problems are discussed.

\subsection{Sources of PU Data}
PU data naturally arises in the following settings.

An \emph{automatic diagnosis} system aims to predict if a patient has a disease. The data for such a system would consist of patients that were diagnosed with the disease and patients that were not. However, not being diagnosed is not equal to not having it. Many diseases, like diabetes, often go undiagnosed~\cite{Claesen2015BuildingCT}. Diagnoses patients are thus positive examples, while undiagnosed are unlabeled.

Sometimes, \emph{positive examples are easier to obtain}. Recommendation systems, for example, can use previous purchases or likes as examples for items of interest. Similarly, some spam mails will be tagged as such. Purchased or tagged items are thus positive examples, while the others are unlabeled.

\emph{Indirect labels} can be used to get some labeled examples. For example, to classify active students based on university records, the students that are registered in university sport classes are active. Other students are unlabeled.

The \emph{case-control} scenario comes from the setting where two datasets are used and one is known to only have positive examples. For example, to predict one's socioeconomic status from her health record, positive examples could be gathered from health centers in upper-class neighborhoods and unlabeled examples from a random selection of health centers.

\emph{Negative-class dataset shift} occurs when the distribution of the negative examples changes while the positive distribution remains the same. This happens, for example, in adversarial scenarios. In this case it might be easier to obtain a new representative sample from the entire distribution than to label characteristic examples from the new negative distribution~\cite{Plessis2015ConvexFF}.

In surveys, \emph{under-reporting} occurs when participants are likely to give false negative responses~\cite{Sechidis2017DealingWU}. This occurs for issues that have social stigma, such as maternal smoking. Research has shown that smoking may be underestimated by up to 47\%~\cite{Gorber2009TheAO}. In this setting, a negative response is really an unlabeled example.

The goal of \emph{one-class classification} is to recognize examples from the class of interest, i.e., the  positive class, from the entire population. When an unlabeled dataset is available that represents the entire population, then this can be seen as learning from positive and unlabeled data~\cite{khan_madden_2014}. In this case, the negative class often has a large variety, for which it is difficult to label a representative sample~\cite{Li2011APA}.

\emph{Inlier-based outlier detection} has access to a representative sample of inliers, in addition to the standard unsupervised data. With this information, more powerful outlier detection is possible~\cite{Hido2008InlierBasedOD,Smola2009RelativeND}. This task can be phrased as PU learning, with the inliers as the positive class~\cite{blanchard2010semi}.

Automatic \emph{knowledge base completion} is inherently a positive and unlabeled problem. Automatically constructed knowledge bases are necessarily incomplete and only contain true facts~\cite{Galarraga2015,neelakantan2015compositional}. The unlabeled examples are the facts that are considered to be added to the knowledge base.

\emph{Identification} problems aim to identify examples in an unlabeled dataset that are similar to the provided examples. For example, disease gene identification aims to identify new disease-genes~\cite{Mordelet2011ProDiGePO}.

\subsection{Applications}

PU learning has been applied to a variety of problems. 

\emph{Disease gene identification} aims to identify which genes from the human genome are causative for diseases. Here, all the known disease genes are positive examples, while all other candidates, that can be generated by traditional linkage analysis, genes are unlabeled. To check all of the candidates individually would be very costly. With PU learning, a promising subset can be discovered. Several PU methods were developed to this end: ProDiGe is a method based on bagging SVMs~\cite{Mordelet2011ProDiGePO,mordelet2014bagging}, PUDI is also a weighted SVM method, but they have different weights for four identified groups of unlabeled examples: reliable negative, likely positive, likely negative and weakly negative~\cite{Yang2012PositiveunlabeledLF}, EPU uses  multiple biological data sources and trains an ensemble model on those~\cite{Yang2014EnsemblePU}.

\emph{Protein complexes} are a set of interacting proteins for specific biological activities. Such complexes can be predicted as subgraphs from protein-protein interaction networks. Known complexes are positive examples and all other possibilities are unlabeled. This problem has been addressed using a non-traditional classifier approach~\cite{elkan2008learning,Zhao2016Protein}.

A \emph{gene regulatory network} is a set of interacting genes that control cell functions. Using the non-traditional classifier method with SVMs, the relationships between activation profiles of gene pairs can be identified~\cite{elkan2008learning,cerulo2010learning}. Bagging SVMs have been employed to identify which genes are under control of which transcription factors~\cite{mordelet2014bagging,Mordelet2013SupervisedIO}.

In the field of \emph{drug discovery}, the tasks of \emph{drug repositioning}, which looks for interactions between drugs and diseases, and \emph{drug-drug-interactions} are very important. To find these interactions, a pairwise scoring function can be trained so that known interactions score higher than pairs which are not known to interact~\cite{Liu2017ComputationalDD}. The rationale behind this method is similar to RSVM~\cite{Sellamanickam2011APR}.

\emph{Ecological modeling of the habitat of species} aims to model where certain animals appear. An observed animal at a certain location provides positive examples. However not observing an animal does not mean that  it never comes there. An EM algorithm on top of logistic regression that finds the optimal likelihood model, given the class prior, was proposed to address this application~\cite{ward2009presence}.

The goal of \emph{targeted marketing} is to only promote products to potential buyers. The difficulty is to identify these customers. A biased SVM approach has been used to identify heat pump owners based on smart meter data, prior sales and weather data~\cite{liu2003building,Fei2013HeatPD}. For online retail, purchase data is often used as positive examples. However, for durable goods, like televisions, only a small fraction of potential customers will purchase it, not because they are not interested, but because already have one or are waiting for the right time, etc. A custom algorithm was developed for this application~\cite{Yi2017ScalableDR}.

\emph{Remote sensing} data, like satellite pictures, can be used to classify certain areas. While examples can be given for the class of interest, it can be hard to identify negative examples, because those are too diverse to be labeled. A non-traditional classifier can be used in such a context~\cite{elkan2008learning,Li2011APA}.

Local descriptors play an important role in \emph{localization} of, for example, mobile robots from laser scanner data. However, in some natural environment, many of the local descriptors might be unreliable and are better filtered out than used. To this end, the non-traditional random forest can be used, where the unlabeled examples are subsampled in a similar way as for bagging SVMs~\cite{elkan2008learning,mordelet2014bagging,Breiman2001,Latulippe2013AcceleratedRP}.

\emph{Recommender systems} can suffer from deceptive reviews, which are dishonest positive or negative reviews. These reviews should therefore be filtered out. Some positive examples of such reviews can be provided, but all other reviews to be checked are unlabeled~\cite{ren2014positive}. 

\emph{Focused web crawlers} search for relevant web pages given a query. Such a web crawler chooses to follow a link or not, based on the link's context. It is much easier to provide positive examples of such contexts than to provide a good sample of negative examples.  Therefore the WVC and PSOC methods have been used to address this problem~\cite{Peng2007SVMBA}.

\revisions{
In \emph{time series anomaly detection}, the goal is to identify portions of the data characterized by presence of unexpected or abnormal behavior. In the case of water usage data~\cite{vercruyssen2018semi}, recognizing certain patterns can play an important role in an anomaly detector. Because it is too time consuming to annotate all pattern occurrences in the data, an expert will typically annotate a few segments containing the pattern. The task of identifying the remaining patterns~\cite{vercrussyen2020} can be viewed as a PU problem with the annotated segments serving as positive examples and unannotated segments as unlabeled examples, as these may or may  not contain the pattern. The inductive bagging SVM~\cite{mordelet2014bagging} has been shown to work well for this task.}

\section{Related Fields}
\label{sec:reldom}
This section briefly discusses the fields that are closely related to PU learning.

\subsection{Semi-Supervised Learning}
The goal of semi-supervised learning is to learn from labeled and unlabeled data~\cite{chapelle2009semi}. In contrast to PU learning, labeled examples of all classes are assumed to be present in the data. Also, semi-supervised learning can go beyond binary classification tasks.  Although semi-supervised methods cannot be applied directly to PU learning, some approaches have been ported from one domain to the other~\cite{Denis2003TextCA,pelckmans2009transductively}.

For semi-supervised learning methods that incorporate the class prior, it is usually assumed that the class prior can be readily estimated from the labeled data, i.e., that positive and negative examples are selected to be labeled with the same probability. However, recently a matching method has been proposed to estimate the class prior when this is not the case~\cite{Plessis2012SemiSupervisedLO}.

\subsection{One-Class Classification}
The goal of one-class classification is to learn a model that identifies examples from a certain class: the positive class, when only examples of that class are available~\cite{khan_madden_2014}. It can be seen as training a binary classifier where the negative class consists of all other possible classes. This is in contrast to PU learning, where the domain of interest is defined by the unlabeled data. Also, the unlabeled data enables finding low-density areas which are likely to be classification boundaries under the separability assumption. Under the SCAR assumption, areas with relatively more unlabeled examples than positive ones indicate a negative region, which would not be clear with only positive examples.

\subsection{Classification in the Presence of Label Noise}
Label noise occurs when some of the class labels in the data are erroneous, i.e., when some examples have a class label that does not correspond with its true class value. A common interpretation of PU learning is that it is the specific type of label noise\revisions{, called \emph{one-sided label noise,} where the positive examples can be incorrectly labeled as negative~\cite{Scott2013ClassificationWA}}. All the biased learning methods are based on this interpretation.

Just like the SCAR assumption was proposed in analogy with the MCAR assumption from missing data, a taxonomy for mislabeling mechanisms was proposed in analogy with the missing data taxonomy~\cite{frenay2014classification}:
\begin{description}
	\item [\textit{NCAR}] \emph{Noisy Completely At Random} Every class label has exactly the same probability to be erroneous, independent of the attribute values of the example or the true class value.
	\item [\textit{NAR}] \emph{Noisy At Random} The probability for a class label to be erroneous depends completely on the true class value\revisions{, this is also known as asymmetric label noise}.
	\item [\textit{NNAR}] \emph{Noisy Not At Random} The probability for a class label to be erroneous depends on the attribute values
\end{description}
The SCAR labeling mechanism corresponds to the NAR mislabeling mechanism, where the mislabeling probability for the positive and negative class are $1-c$ and $0$ respectively. \revisions{The label noise literature refers to mislabeling probability $1-c$ as the noise rate or flip rate $\rho_{+1}$~\cite{Scott2013ClassificationWA,Natarajan2013LearningWN}}.

\revisions{Because SCAR PU Learning is a specific setting of learning with NAR noisy labels, the SCAR methods can often be generalized to NAR. For example, rebalancing methods, where the instances get class-dependent weights, and empirical-risk-minimization based methods both exists for learning with NAR noisy labels~\cite{Natarajan2013LearningWN,Natarajan2017CostSensitiveLW}. Rank pruning was also proposed for the general NAR noisy labels setting~\cite{Northcutt2017LearningWC}.}

\subsection{Missing Data}
When working with missing data, the missingness mechanism that dictates which values are missing plays a crucial role, just like the labeling mechanism for PU learning. The missingness mechanisms are generally divided into three classes~\cite{rubin1976inference,little2002statistical}:

\begin{description}
	\item [\textit{MCAR}] \emph{Missing Completely At Random} Every attribute has exactly the same probability to be missing, independent of the other attribute values of the example and the value of the missing attribute.
	\item [\textit{MAR}] \emph{Missing At Random} The probability for an attribute to be missing depends completely on the observable attributes of the example.
	\item [\textit{MNAR}] \emph{Missing Not At Random} The probability for an attribute to be missing depends on the value that is missing.
\end{description}

The SCAR and SAR assumptions were introduced in analogy with MCAR and MAR. However, it is important to note that within the missing data taxonomy, SCAR and SAR actually both belong to the MNAR class, because positive and negative class values have a different probabilities to be missing: $c$ or $e(x)$ and $0$ respectively. The class values are missing (completely) at random only if just the population of positive examples is considered.  Moreno et al. (2012) proposed a new missingness class: \emph{Missing Completely At Random-Class Dependent (MAR-C)}, where per class, the data is MCAR, as is the case for SCAR.

\subsection{Multiple-Instance Learning}

The goal of multiple-instance learning is to train a binary classifier. Instead of positive and negative examples, the learner is provided with bags that are labeled positive if at least one of the examples in the bag is positive and negative otherwise. This setting can be phrased as PU learning, or actually NU learning, as the classes are switched. All the examples in a negative bag are known to be negative and can therefore get a negative label, while examples in a positive bag can be both positive and negative and therefore are considered unlabeled. Following this insight, classifiers from either domains can be used to solve the task of the other domain~\cite{Li2013TheLB}.

\section{Conclusions and Perspectives}

PU learning is a very active area of research within the machine learning community. We will end by tying the survey back to the central PU learning research questions and discussing key future directions. 

\subsection{Questions Revisited}
At the end of the introduction, we posed seven research questions frequently addressed in PU learning research.  To conclude, we will revisit these questions and try to synthesize answers to each one. 

\paragraph{How can we formalize the problem of learning from PU data?}
The PU learning literature always assumes one of two learning scenarios: single-training-set or case-control, which are discussed in Section~\ref{sec:prelim}. The former assumes one dataset that is an i.i.d. sample of the true distribution. A subset of the positive examples of the dataset are labeled while the remaining examples are unlabeled. The latter scenario assumes two independently drawn datasets: an i.i.d. sample of the true distribution (unlabeled) and a sample of the positive part of the true distribution (positive). The labeled examples are selected from the positive subset or the positive distribution according to the labeling mechanism.

\paragraph{What assumptions are typically made about PU data in order to facilitate
the design of learning algorithms?}
As discussed in Section~\ref{sec:assumptions}, assumptions are needed either about the data distribution, or the labeling mechanism, or both.  The most common assumptions about the data distribution are separable classes and smoothness, which form the basis for the two-step learning techniques. The most common labeling mechanism assumption is selected completely at random (SCAR) assumption, where postures that the set of labeled examples is a uniformly random subset of the positive examples. It greatly simplifies learning and it serves as the basis of all class-prior based methods. Recently, the more realistic SAR assumption has been proposed which assumes that the labeling mechanism depends on the attributes.

\paragraph{Can we estimate the class prior from PU data and why is this useful?}
By making assumptions about the data and/or labeling mechanism it is possible to estimate the label frequency and hence class prior in certain conditions (Section~\ref{sec:assumptions_alpha}). Multiple different techniques have been proposed for this task (Section~\ref{sec:back_alphaest}). The power and usefulness of this piece of information is that facilitates the design of algorithms for learning from PU data (Section~\ref{sec:classpriorincorp}). This is effectively done by estimating the expected number of positive and negative examples of the data, which can be accomplished by either weighting the data and then applying standard algorithms or directly modifying algorithms to work with fractional counts. 

\paragraph{How can we learn a model from PU data?}
Section~\ref{sec:pumethods} shows that most PU learning methods belong to one of three categories: two-step techniques, biased learning and class prior incorporation methods. Two-step techniques begin by identifying reliable negative (and sometimes positive) examples and then using the labeled and reliable examples to train a classifier. The biased methods treat the unlabeled examples as belonging to the negative class, but attribute a larger loss to false positives than false negatives. Class prior incorporation methods use the class prior to weight the unlabeled data or modify machine learning algorithms to reason about the expected number of positive and negative examples in the unlabeled data.

\paragraph{How can we evaluate models in a PU setting?}
This is an area that has perhaps received less attention in the literature. This can be approached in two general ways, both of which exploit the SCAR assumption. One is to use the (estimated) class prior and construction bounds for traditional evaluation metrics such as accuracy. The other is to design metrics that can be computed based on the observed information (e.g., could be computed using only positive examples) which are proxies for standard metrics. This was discussed in Section~\ref{sec:evalmetrics}.

\paragraph{When and why does PU data arise in real-world applications?}
As outlined in Section~\ref{sec:sourcesandapplications}, PU data arises in many different fields. At a high-level, it occurs in the following types of situations:
\begin{enumerate}
    \item When only "positive" information is recorded such as in an electronic medical record or a knowledge base that stores facts, where the absence of information does not imply something is not true; 
    \item People have a reason to be deceptive and not report such as lying about smoking when pregnant in a survey or an athlete hiding an injury in order to keep playing;
    \item Where it is much easier to identify one class than another, such as certain bioinformatics problems or remote sensing. 
\end{enumerate}

\paragraph{How does PU learning relate to other areas of machine learning}
Section~\ref{sec:reldom} shows that PU learning is related to numerous areas of machine learning. Most obviously, it is a special case of standard semi-supervised learning. The key differences are that typically semi-supervised approaches have access to at least some examples of all classes, and that semi-supervised approaches go beyond binary classification tasks. Similarly, it can also be viewed through the prism of learning with label noise. Again, PU learning is a specialization in that corresponds to one type of noise: that where positive examples are possibly incorrectly labeled as negative. Some of the nomenclature about labeling mechanisms has been inspired by the long standing field of working with missing data. Finally, it also tied to one-class classification, learning with missing data and multiple-instance learning.

\subsection{Future Directions}
Given that PU data naturally arises in many real-world datasets, it should continue to be an active area of machine learning research. The key open questions will revolve around making sure the assumptions and settings considered within PU learning align with real-world PU tasks. Therefore, there are several key directions that PU could take, which we now expand upon.

\revisions{
\paragraph{More realistic labeling mechanisms and corresponding learning methods}~}One important area of research is to consider more realistic assumptions about the labeling mechanism. Until this year, the vast majority of work had focused on the SCAR assumption, given that it facilitates analysis. \revisions{However, this assumption clearly often does not hold in practice. On the other side of the spectrum, there is the SAR assumption, which is so general that it essentially always holds. However, it is so general that effective learning in this setting requires making additional assumptions. The probabilistic gap assumption finds some middle-ground. 
However, it does not always apply. For example, a professional sports player (e.g., a football or soccer player) in a contract year may be less likely to report a minor injury, but this has no relationship with the probability of a player getting injured.}  Therefore, researchers should continue to consider how to formalize different labeling assumptions that more closely resemble how PU data naturally arises within real-world applications. 
\revisions{
Additionally, learning methods should be developed that leverage these labeling assumptions.
}

\paragraph{An empirical comparison of PU learning approaches} As this survey shows, a wide variety of PU learning approaches have been proposed. While many of the approaches have a strong theoretical basis, presuming certain assumptions hold, we still lack a complete empirical understanding of how the various approaches perform. In the literature, papers typically compare a hand full of approaches on a small number of datasets (i.e., often less than ten). Moreover, the considered datasets vary by paper. An extensive evaluation could help provide us with more insight into which methods are preferred and which assumptions are reasonable for obtaining good performance in practice.

\revisions{
\paragraph{Evaluating classifier performance on PU data} The standard approach to evaluating a PU classifier's generalization ability is  to assume a fully labeled test set. While this is convenient, it does not conform to the motivation of learning from PU data. There has been some work on evaluating classifier performance using PU data, which is a more challenging setting. However, much of this work is theoretical, and there has been little (if any) direct quantitative comparison among the various approaches (e.g.,~\cite{claesen2015assessing,Jain2017RecoveringTC,Sechidis2014StatisticalHT}). An important future direction is understanding how these metrics perform in practice. Furthermore, often these approaches rely on the SCAR assumption (e.g.,~\cite{claesen2015assessing,Jain2017RecoveringTC}) and it will be important to design metrics that work for other labeling mechanisms. 
}

\paragraph{Real-world PU benchmarks} The current evaluation paradigm largely consists of using existing, fully labeled datasets and converting them into a PU setting. This has advantages and disadvantages. The positive aspect is it provides a controlled manner in which to assess performance. This setup typically ensures that the assumptions made in the paper are respected. The disadvantage is that we then lack an understanding about what will happen "in the wild" when the assumptions are violated. One partial remedy would be to encourage authors to simulate these violations. Ideally, several real-world PU benchmarks could be created and released, which would greatly benefit the community. We do note that in the fully PU setting, evaluation would be very tricky. One promising domain for this is knowledge base completion. While this is often not view through the lens of PU learning, the task certainly could be categorized in this way. 

\paragraph{PU learning in relational domains} The vast majority of PU learning work has focused on the propositional setting. There has been a renewed interest recently in learning from relational data. This dovetails with the previous suggestion in that knowledge base completion is inherently a relational problem. Therefore, it may be fruitful to further explore how to enable PU learning in relational domains both from a theoretical and algorithmic perspective. 

\begin{acknowledgements}
JB is supported by IWT (SB/141744).  JD is partially supported by FWO-Vlaanderen (G0D8819N), KU Leuven Research Fund (C14/17/070), and the Flemish Government under the “Onderzoeksprogramma Artifici{\"e}le Intelligentie (AI) Vlaanderen” programme.
\end{acknowledgements}
\appendix

\bibliographystyle{spmpsci} 
\bibliography{allpapers}  

\begin{thebibliography}{100}
\providecommand{\url}[1]{{#1}}
\providecommand{\urlprefix}{URL }
\expandafter\ifx\csname urlstyle\endcsname\relax
  \providecommand{\doi}[1]{DOI~\discretionary{}{}{}#1}\else
  \providecommand{\doi}{DOI~\discretionary{}{}{}\begingroup
  \urlstyle{rm}\Url}\fi

\bibitem{basile2017de}
Basile, T.M., Di~Mauro, N., Esposito, F., Ferilli, S., Vergari, A.: Density
  estimators for positive-unlabeled learning.
\newblock In: New Frontiers in Mining Complex Patterns: 6th International
  Workshop, NFMCP 2017, Held in Conjunction with ECML-PKDD 2017, Skopje,
  Macedonia, September 18-22, 2017, Revised Selected Papers, vol. 10785, pp.
  49--64. Springer (2018)

\bibitem{bekker2018aaai}
Bekker, J., Davis, J.: Estimating the class prior in positive and unlabeled
  data through decision tree induction.
\newblock In: Proceedings of the 32th AAAI Conference on Artificial
  Intelligence, pp. 2712--2719 (2018)

\bibitem{bekkerilp17}
Bekker, J., Davis, J.: Positive and unlabeled relational classification through
  label frequency estimation.
\newblock In: N.~Lachiche, C.~Vrain (eds.) Inductive Logic Programming, pp.
  16--30. Springer International Publishing, Cham (2018)

\bibitem{bekker2019ecml}
Bekker, J., Robberechts, P., Davis, J.: Beyond the selected completely at
  random assumption for learning from positive and unlabeled data.
\newblock In: ECML PKDD: Joint European Conference on Machine Learning and
  Knowledge Discovery in Databases. Springer (2019)

\bibitem{blanchard2010semi}
Blanchard, G., Lee, G., Scott, C.: Semi-supervised novelty detection.
\newblock Journal of Machine Learning Research \textbf{11}, 2973--3009 (2010)

\bibitem{Blockeel2017PUlearningDC}
Blockeel, H.: Pu-learning disjunctive concepts in ilp.
\newblock In: ILP 2017 late breaking papers (2017)

\bibitem{Blum1998CombiningLA}
Blum, A., Mitchell, T.: Combining labeled and unlabeled data with co-training.
\newblock In: Proceedings of the eleventh annual conference on Computational
  learning theory, pp. 92--100. ACM (1998)

\bibitem{Breiman2001}
Breiman, L.: Random forests.
\newblock Machine Learning \textbf{45}(1), 5--32 (2001).
\newblock \doi{10.1023/A:1010933404324}.
\newblock \urlprefix\url{https://doi.org/10.1023/A:1010933404324}

\bibitem{Calvo2007LBC}
Calvo, B., Larra\~{n}aga, P., Lozano, J.A.: Learning bayesian classifiers from
  positive and unlabeled examples.
\newblock Pattern Recogn. Lett. \textbf{28}(16), 2375--2384 (2007).
\newblock \doi{10.1016/j.patrec.2007.08.003}.
\newblock \urlprefix\url{http://dx.doi.org/10.1016/j.patrec.2007.08.003}

\bibitem{cerulo2010learning}
Cerulo, L., Elkan, C., Ceccarelli, M.: Learning gene regulatory networks from
  only positive and unlabeled data.
\newblock BMC bioinformatics \textbf{11}(1), 228 (2010)

\bibitem{Chang2016PositiveUnlabeledLI}
Chang, S., Zhang, Y., Tang, J., Yin, D., Chang, Y., Hasegawa-Johnson, M.A.,
  Huang, T.S.: Positive-unlabeled learning in streaming networks.
\newblock In: Proceedings of the 22nd ACM SIGKDD International Conference on
  Knowledge Discovery and Data Mining, pp. 755--764. ACM (2016)

\bibitem{chapelle2009semi}
Chapelle, O., Scholkopf, B., Zien, A.: Semi-supervised learning.
\newblock IEEE Transactions on Neural Networks \textbf{20}(3), 542--542 (2009)

\bibitem{Chaudhari2012LPU}
Chaudhari, S., Shevade, S.: Learning from positive and unlabelled examples
  using maximum margin clustering.
\newblock In: Proceedings of the 19th International Conference on Neural
  Information Processing, vol.~3, pp. 465--473. Springer-Verlag, Berlin,
  Heidelberg (2012)

\bibitem{Chiaroni2018LearningWA}
Chiaroni, F., Rahal, M.C., Hueber, N., Dufaux, F.: Learning with a generative
  adversarial network from a positive unlabeled dataset for image
  classification.
\newblock In: IEEE International Conference on Image Processing (2018)

\bibitem{claesen2015assessing}
Claesen, M., Davis, J., De~Smet, F., De~Moor, B.: Assessing binary classifiers
  using only positive and unlabeled data.
\newblock arXiv preprint arXiv:1504.06837  (2015)

\bibitem{claesen2015building}
Claesen, M., De~Smet, F., Gillard, P., Mathieu, C., De~Moor, B.: Building
  classifiers to predict the start of glucose-lowering pharmacotherapy using
  belgian health expenditure data.
\newblock arXiv preprint arXiv:1504.07389  (2015)

\bibitem{Claesen2015BuildingCT}
Claesen, M., Smet, F.D., Gillard, P., Mathieu, C., Moor, B.D.: Building
  classifiers to predict the start of glucose-lowering pharmacotherapy using
  belgian health expenditure data.
\newblock CoRR \textbf{abs/1504.07389} (2015)

\bibitem{Claesen2015ARE}
Claesen, M., Smet, F.D., Suykens, J.A.K., Moor, B.D.: A robust ensemble
  approach to learn from positive and unlabeled data using svm base models.
\newblock Neurocomputing \textbf{160}, 73--84 (2015)

\bibitem{claesen2015robust}
Claesen, M., Smet, F.D., Suykens, J.A.K., Moor, B.D.: A robust ensemble
  approach to learn from positive and unlabeled data using svm base models.
\newblock Neurocomputing \textbf{160}, 73--84 (2015)

\bibitem{denis2005learning}
Denis, F., Gilleron, R., Letouzey, F.: Learning from positive and unlabeled
  examples.
\newblock Theoretical Computer Science \textbf{348}(1), 70--83 (2005)

\bibitem{Denis2003TextCA}
Denis, F., Laurent, A., Gilleron, R., Tommasi, M.: Text classification and
  co-training from positive and unlabeled examples.
\newblock In: Proceedings of the ICML 2003 workshop: the continuum from labeled
  to unlabeled data, pp. 80--87 (2003)

\bibitem{Plessis2015ConvexFF}
Du~Plessis, M., Niu, G., Sugiyama, M.: Convex formulation for learning from
  positive and unlabeled data.
\newblock In: International Conference on Machine Learning, pp. 1386--1394
  (2015)

\bibitem{Plessis2014ClassPE}
Du~Plessis, M.C., Sugiyama, M.: Class prior estimation from positive and
  unlabeled data.
\newblock IEICE TRANSACTIONS on Information and Systems \textbf{97}(5),
  1358--1362 (2014)

\bibitem{Elkan2001TheFO}
Elkan, C.: The foundations of cost-sensitive learning.
\newblock In: Proceedings of the seventeenth international joint conference on
  artificial intelligence, vol.~17, pp. 973--978. Lawrence Erlbaum Associates
  Ltd (2001)

\bibitem{elkan2008learning}
Elkan, C., Noto, K.: Learning classifiers from only positive and unlabeled
  data.
\newblock In: Proceedings of the 14th ACM SIGKDD international conference on
  Knowledge discovery and data mining, pp. 213--220. ACM (2008)

\bibitem{Fei2013HeatPD}
Fei, H., Kim, Y., Sahu, S., Naphade, M., Mamidipalli, S.K., Hutchinson, J.:
  Heat pump detection from coarse grained smart meter data with positive and
  unlabeled learning.
\newblock In: Proceedings of the 19th ACM SIGKDD international conference on
  Knowledge discovery and data mining, pp. 1330--1338. ACM (2013)

\bibitem{frenay2014classification}
Fr{\'e}nay, B., Verleysen, M.: Classification in the presence of label noise: a
  survey.
\newblock IEEE transactions on neural networks and learning systems
  \textbf{25}(5), 845--869 (2014)

\bibitem{Fung2006TextCW}
Fung, G.P.C., Yu, J.X., Lu, H., Yu, P.S.: Text classification without negative
  examples revisit.
\newblock IEEE Transactions on Knowledge and Data Engineering \textbf{18},
  6--20 (2006)

\bibitem{Galarraga2015}
Gal{\'a}rraga, L., Teflioudi, C., Hose, K., Suchanek, F.M.: Fast rule mining in
  ontological knowledge bases with {AMIE+}.
\newblock The VLDB Journal—The International Journal on Very Large Data Bases
  \textbf{24}(6), 707--730 (2015)

\bibitem{Gan2017BBN}
Gan, H., Zhang, Y., Song, Q.: Bayesian belief network for positive unlabeled
  learning with uncertainty.
\newblock Pattern Recogn. Lett. \textbf{90}(C), 28--35 (2017).
\newblock \doi{10.1016/j.patrec.2017.03.007}.
\newblock \urlprefix\url{https://doi.org/10.1016/j.patrec.2017.03.007}

\bibitem{Gorber2009TheAO}
Gorber, S.C., Schofield-Hurwitz, S., Hardt, J.S., Levasseur, G., Tremblay,
  M.D.: The accuracy of self-reported smoking: a systematic review of the
  relationship between self-reported and cotinine-assessed smoking status.
\newblock Nicotine \& tobacco research : official journal of the Society for
  Research on Nicotine and Tobacco \textbf{11}(1), 12--24 (2009)

\bibitem{he2018idpu}
He, F., Liu, T., Webb, G.I., Tao, D.: Instance-dependent pu learning by
  bayesian optimal relabeling.
\newblock arXiv preprint arXiv:1808.02180  (2018)

\bibitem{He2010NaiveBC}
He, J., Zhang, Y., Li, X., Wang, Y.: Naive bayes classifier for positive
  unlabeled learning with uncertainty.
\newblock In: Proceedings of the 2010 SIAM International Conference on Data
  Mining, pp. 361--372. SIAM (2010)

\bibitem{He2011BCP}
He, J., Zhang, Y., Li, X., Wang, Y.: Bayesian classifiers for positive
  unlabeled learning.
\newblock In: Proceedings of the 12th International Conference on Web-age
  Information Management, WAIM'11, pp. 81--93. Springer-Verlag, Berlin,
  Heidelberg (2011).
\newblock \urlprefix\url{http://dl.acm.org/citation.cfm?id=2035562.2035574}

\bibitem{Hido2008InlierBasedOD}
Hido, S., Tsuboi, Y., Kashima, H., Sugiyama, M., Kanamori, T.: Inlier-based
  outlier detection via direct density ratio estimation.
\newblock 2008 Eighth IEEE International Conference on Data Mining pp. 223--232
  (2008)

\bibitem{Hou2018GenerativeAP}
Hou, M., Chaib-draa, B., Li, C., Zhao, Q.: Generative adversarial
  positive-unlabelled learning.
\newblock In: Proceedings of the Twenty-Seventh International Joint Conference
  on Artificial Intelligence, {IJCAI-18}, pp. 2255--2261 (2018).
\newblock \doi{10.24963/ijcai.2018/312}

\bibitem{hsieh2015pu}
Hsieh, C.J., Natarajan, N., Dhillon, I.: P{U} learning for matrix completion.
\newblock In: International Conference on Machine Learning, pp. 2445--2453
  (2015)

\bibitem{Ienco2016PUL}
Ienco, D., Pensa, R.G.: Positive and unlabeled learning in categorical data.
\newblock Neurocomput. \textbf{196}(C), 113--124 (2016).
\newblock \doi{10.1016/j.neucom.2016.01.089}.
\newblock \urlprefix\url{https://doi.org/10.1016/j.neucom.2016.01.089}

\bibitem{Ienco2012FromCT}
Ienco, D., Pensa, R.G., Meo, R.: From context to distance: Learning
  dissimilarity for categorical data clustering.
\newblock ACM Transactions on Knowledge Discovery from Data (TKDD)
  \textbf{6}(1), 1--25 (2012)

\bibitem{Jain2016EstimatingTC}
Jain, S., White, M., Radivojac, P.: Estimating the class prior and posterior
  from noisy positives and unlabeled data.
\newblock In: Advances in Neural Information Processing Systems, pp. 2693--2701
  (2016)

\bibitem{Jain2017RecoveringTC}
Jain, S., White, M., Radivojac, P.: Recovering true classifier performance in
  positive-unlabeled learning.
\newblock In: Proceedings of the 31st AAAI Conference on Artificial
  Intelligence, pp. 2066--2073 (2017)

\bibitem{Jain2016NonparametricSL}
Jain, S., White, M., Trosset, M.W., Radivojac, P.: Nonparametric
  semi-supervised learning of class proportions.
\newblock arXiv preprint arXiv:1601.01944  (2016)

\bibitem{Jiang2009ANB}
Jiang, L., Zhang, H., Cai, Z.: A novel bayes model: Hidden naive bayes.
\newblock IEEE Transactions on knowledge and data engineering \textbf{21}(10),
  1361--1371 (2009)

\bibitem{Ke2017GlobalAL}
Ke, T., Jing, L., Lv, H., Zhang, L., Hu, Y.: Global and local learning from
  positive and unlabeled examples.
\newblock Applied Intelligence \textbf{48}, 2373--2392 (2017)

\bibitem{KE2018422}
Ke, T., Lv, H., Sun, M., Zhang, L.: A biased least squares support vector
  machine based on {M}ahalanobis distance for {PU} learning.
\newblock Physica A: Statistical Mechanics and its Applications \textbf{509},
  422 -- 438 (2018).
\newblock \doi{https://doi.org/10.1016/j.physa.2018.05.128}.
\newblock
  \urlprefix\url{http://www.sciencedirect.com/science/article/pii/S0378437118306794}

\bibitem{ke2012building}
Ke, T., Yang, B., Zhen, L., Tan, J., Li, Y., Jing, L.: Building
  high-performance classifiers using positive and unlabeled examples for text
  classification.
\newblock In: International Symposium on Neural Networks, pp. 187--195.
  Springer (2012)

\bibitem{khan_madden_2014}
Khan, S., Madden, M.: One-class classification: taxonomy of study and review of
  techniques.
\newblock The Knowledge Engineering Review  (2014)

\bibitem{Khot2014RelationalOC}
Khot, T., Natarajan, S., Shavlik, J.W.: Relational one-class classification: A
  non-parametric approach.
\newblock In: Proceedings of the 28th AAAI Conference on Artificial
  Intelligence, pp. 2453--2460 (2014)

\bibitem{Kiryo2017PositiveUnlabeledLW}
Kiryo, R., Niu, G., du~Plessis, M.C., Sugiyama, M.: Positive-unlabeled learning
  with non-negative risk estimator.
\newblock In: Advances in Neural Information Processing Systems, pp. 1675--1685
  (2017)

\bibitem{Kull2017BetaCA}
Kull, M., de~Menezes~e Silva~Filho, T., Flach, P.A.: Beta calibration: a
  well-founded and easily implemented improvement on logistic calibration for
  binary classifiers.
\newblock In: Proceedings of the twentieth International Conference on
  Artificial Intelligence and Statistics, pp. 623--631 (2017)

\bibitem{Latulippe2013AcceleratedRP}
Latulippe, M., Drouin, A., Giguere, P., Laviolette, F.: Accelerated robust
  point cloud registration in natural environments through positive and
  unlabeled learning.
\newblock In: Proceedings of the 23th International Joint Conference on
  Artifical Intelligence, pp. 2480--2487 (2013)

\bibitem{lee2003learning}
Lee, W.S., Liu, B.: Learning with positive and unlabeled examples using
  weighted logistic regression.
\newblock In: Proceedings of the Twentieth International Conference on Machine
  Learning, pp. 448--455 (2003)

\bibitem{Li2011APA}
Li, W., Guo, Q., Elkan, C.: A positive and unlabeled learning algorithm for
  one-class classification of remote-sensing data.
\newblock IEEE Transactions on Geoscience and Remote Sensing \textbf{49},
  717--725 (2011)

\bibitem{li2003learning}
Li, X., Liu, B.: Learning to classify texts using positive and unlabeled data.
\newblock In: Proceedings of the eighteenth International Joint Conference on
  Artifical Intelligence, vol.~3, pp. 587--592 (2003)

\bibitem{Li2007LearningTI}
Li, X., Liu, B., Ng, S.K.: Learning to identify unexpected instances in the
  test set.
\newblock In: Proceedings of the 20th International Joint Conference on
  Artifical Intelligence, vol.~7, pp. 2802--2807 (2007)

\bibitem{Li2005LearningFP}
Li, X.L., Liu, B.: Learning from positive and unlabeled examples with different
  data distributions.
\newblock In: European Conference on Machine Learning, pp. 218--229. Springer
  (2005)

\bibitem{Li2010NegativeTD}
Li, X.L., Liu, B., Ng, S.K.: Negative training data can be harmful to text
  classification.
\newblock In: Proceedings of the 2010 conference on empirical methods in
  natural language processing, pp. 218--228. Association for Computational
  Linguistics (2010)

\bibitem{li2009positive}
Li, X.L., Yu, P.S., Liu, B., Ng, S.K.: Positive unlabeled learning for data
  stream classification.
\newblock In: Proceedings of the 2009 SIAM International Conference on Data
  Mining, pp. 259--270. SIAM (2009)

\bibitem{Li2013TheLB}
Li, Y., Tax, D.M., Duin, R.P., Loog, M.: The link between multiple-instance
  learning and learning from only positive and unlabelled examples.
\newblock In: International Workshop on Multiple Classifier Systems, pp.
  157--166. Springer (2013)

\bibitem{liang2012learning}
Liang, C., Zhang, Y., Shi, P., Hu, Z.: Learning very fast decision tree from
  uncertain data streams with positive and unlabeled samples.
\newblock Information Sciences \textbf{213}, 50--67 (2012)

\bibitem{little2002statistical}
Little, R.J., Rubin, D.B.: Statistical analysis with missing data.
\newblock John Wiley \& Sons (2002)

\bibitem{liu2003building}
Liu, B., Dai, Y., Li, X., Lee, W.S., Yu, P.S.: Building text classifiers using
  positive and unlabeled examples.
\newblock In: Proceedings of the Third IEEE International Conference on Data
  Mining, pp. 179--186. IEEE (2003)

\bibitem{liu2002partially}
Liu, B., Lee, W.S., Yu, P.S., Li, X.: Partially supervised classification of
  text documents.
\newblock In: Proceedings of the Nineteenth International Conference on Machine
  Learning, vol.~2, pp. 387--394. Citeseer (2002)

\bibitem{Liu2014ClusteringbasedMF}
Liu, L., Peng, T.: Clustering-based method for positive and unlabeled text
  categorization enhanced by improved {TFIDF}.
\newblock Journal of Information Science and Engineering \textbf{30},
  1463--1481 (2014)

\bibitem{Liu2016ClassificationWN}
Liu, T., Tao, D.: Classification with noisy labels by importance reweighting.
\newblock IEEE Transactions on Pattern Analysis and Machine Intelligence
  \textbf{38}, 447--461 (2016)

\bibitem{Liu2017ComputationalDD}
Liu, Y., Qiu, S., Zhang, P., Gong, P., Wang, F., Xue, G., Ye, J.: Computational
  drug discovery with dyadic positive-unlabeled learning.
\newblock In: Proceedings of the 2017 SIAM International Conference on Data
  Mining, pp. 45--53. SIAM (2017)

\bibitem{liu2005partially}
Liu, Z., Shi, W., Li, D., Qin, Q.: Partially supervised classification--based
  on weighted unlabeled samples support vector machine.
\newblock In: Proceedings of the International Conference on Advanced Data
  Mining and Applications, pp. 118--129. Springer (2005)

\bibitem{Lu2010SemisupervisedTC}
Lu, F., Bai, Q.: Semi-supervised text categorization with only a few positive
  and unlabeled documents.
\newblock 2010 3rd International Conference on Biomedical Engineering and
  Informatics \textbf{7}, 3075--3079 (2010)

\bibitem{Mahalanobis193649}
Mahalanobis, P.: On the generalised distance in statistics.
\newblock National Institute of Science of India  (1936)

\bibitem{Mordelet2011ProDiGePO}
Mordelet, F., Vert, J.P.: Prodige: Prioritization of disease genes with
  multitask machine learning from positive and unlabeled examples.
\newblock BMC bioinformatics \textbf{12}, 389 (2011)

\bibitem{Mordelet2013SupervisedIO}
Mordelet, F., Vert, J.P.: Supervised inference of gene regulatory networks from
  positive and unlabeled examples.
\newblock Methods in molecular biology \textbf{939}, 47--58 (2013)

\bibitem{mordelet2014bagging}
Mordelet, F., Vert, J.P.: A bagging svm to learn from positive and unlabeled
  examples.
\newblock Pattern Recognition Letters \textbf{37}, 201--209 (2014)

\bibitem{Muggleton:1996}
Muggleton, S.: Learning from positive data.
\newblock In: Selected Papers from the 6th International Workshop on Inductive
  Logic Programming, pp. 358--376 (1996)

\bibitem{Natarajan2013LearningWN}
Natarajan, N., Dhillon, I.S., Ravikumar, P., Tewari, A.: Learning with noisy
  labels.
\newblock In: NIPS (2013)

\bibitem{Natarajan2017CostSensitiveLW}
Natarajan, N., Dhillon, I.S., Ravikumar, P., Tewari, A.: Cost-sensitive
  learning with noisy labels.
\newblock Journal of Machine Learning Research \textbf{18}, 155:1--155:33
  (2017)

\bibitem{natarajan2015pu}
Natarajan, N., Rao, N., Dhillon, I.: P{U} matrix completion with graph
  information.
\newblock In: Computational Advances in Multi-Sensor Adaptive Processing
  (CAMSAP), 2015 IEEE 6th International Workshop on, pp. 37--40. IEEE (2015)

\bibitem{neelakantan2015compositional}
Neelakantan, A., Roth, B., McCallum, A.: Compositional vector space models for
  knowledge base completion.
\newblock In: Proceedings of the 53rd Annual Meeting of the Association for
  Computational Linguistics and the 7th International Joint Conference on
  Natural Language Processing (Volume 1: Long Papers), pp. 156--166.
  Association for Computational Linguistics (2015).
\newblock \doi{10.3115/v1/P15-1016}.
\newblock \urlprefix\url{http://www.aclweb.org/anthology/P15-1016}

\bibitem{nguyen2011positive}
Nguyen, M.N., Li, X.L., Ng, S.K.: Positive unlabeled learning for time series
  classification.
\newblock In: Proceedings of the Seventeenth International Joint Conference on
  Artificial Intelligence, pp. 1421--1426 (2011)

\bibitem{Northcutt2017LearningWC}
Northcutt, C.G., Wu, T., Chuang, I.L.: Learning with confident examples: Rank
  pruning for robust classification with noisy labels.
\newblock In: Proceedings of the Thirty-Third Conference on Uncertainty in
  Artificial Intelligence, UAI'17. AUAI Press (2017).
\newblock \urlprefix\url{http://auai.org/uai2017/proceedings/papers/35.pdf}

\bibitem{pelckmans2009transductively}
Pelckmans, K., Suykens, J.A.: Transductively learning from positive examples
  only.
\newblock In: Proceedings of the European Symposium on Artificial Neural
  Networks, pp. 23--28 (2009)

\bibitem{Peng2007SVMBA}
Peng, T., Zuo, W., He, F.: Svm based adaptive learning method for text
  classification from positive and unlabeled documents.
\newblock Knowledge and Information Systems \textbf{16}, 281--301 (2007)

\bibitem{platt1999probabilistic}
Platt, J., et~al.: Probabilistic outputs for support vector machines and
  comparisons to regularized likelihood methods.
\newblock Advances in large margin classifiers \textbf{10}(3), 61--74 (1999)

\bibitem{Plessis2015ClasspriorEF}
du~Plessis, M., Niu, G., Sugiyama, M.: Class-prior estimation for learning from
  positive and unlabeled data.
\newblock Proceedings of the 7th Asian Conference on Machine Learning pp.
  221--236 (2015)

\bibitem{Plessis2014AnalysisOL}
du~Plessis, M.C., Niu, G., Sugiyama, M.: Analysis of learning from positive and
  unlabeled data.
\newblock In: Advances in Neural Information Processing Systems, pp. 703--711
  (2014)

\bibitem{Plessis2012SemiSupervisedLO}
du~Plessis, M.C., Sugiyama, M.: Semi-supervised learning of class balance under
  class-prior change by distribution matching.
\newblock Neural networks: the Official Journal of the International Neural
  Network Society \textbf{50}, 110--9 (2012)

\bibitem{Qin2012LearningFD}
Qin, X., Zhang, Y., Li, C., Li, X.: Learning from data streams with only
  positive and unlabeled data.
\newblock Journal of Intelligent Information Systems \textbf{40}, 405--430
  (2012)

\bibitem{Ramaswamy2016MixturePE}
Ramaswamy, H., Scott, C., Tewari, A.: Mixture proportion estimation via kernel
  embedding of distributions.
\newblock In: International Conference on Machine Learning, pp. 2052--2060
  (2016)

\bibitem{ren2014positive}
Ren, Y., Ji, D., Zhang, H.: Positive unlabeled learning for deceptive reviews
  detection.
\newblock In: Proceedings of the conference on Empirical Methods in Natural
  Language Processing, pp. 488--498 (2014)

\bibitem{rubin1976inference}
Rubin, D.B.: Inference and missing data.
\newblock Biometrika \textbf{63}(3), 581--592 (1976)

\bibitem{Scott2015ARO}
Scott, C.: A rate of convergence for mixture proportion estimation, with
  application to learning from noisy labels.
\newblock In: Proceedings of The 18th International Conference on Artificial
  Intelligence and Statistics, pp. 838--846 (2015)

\bibitem{scott2009novelty}
Scott, C., Blanchard, G.: Novelty detection: Unlabeled data definitely help.
\newblock In: The 12th International Conference on Artificial Intelligence and
  Statistics, pp. 464--471 (2009)

\bibitem{Scott2013ClassificationWA}
Scott, C., Blanchard, G., Handy, G., Pozzi, S., Flaska, M.: Classification with
  asymmetric label noise: Consistency and maximal denoising.
\newblock In: Conference On Learning Theory (2013)

\bibitem{Sechidis2015MarkovBD}
Sechidis, K., Brown, G.: Markov blanket discovery in positive-unlabelled and
  semi-supervised data.
\newblock In: ECML PKDD: Joint European Conference on Machine Learning and
  Knowledge Discovery in Databases, pp. 351--366. Springer (2015)

\bibitem{Sechidis2017SimpleSF}
Sechidis, K., Brown, G.: Simple strategies for semi-supervised feature
  selection.
\newblock Machine Learning \textbf{107}, 357--395 (2017)

\bibitem{Sechidis2014StatisticalHT}
Sechidis, K., Calvo, B., Brown, G.: Statistical hypothesis testing in positive
  unlabelled data.
\newblock In: ECML PKDD: Joint European Conference on Machine Learning and
  Knowledge Discovery in Databases, pp. 66--81. Springer (2014)

\bibitem{Sechidis2017DealingWU}
Sechidis, K., Sperrin, M., Petherick, E.S., Luj{\'a}n, M., Brown, G.: Dealing
  with under-reported variables: An information theoretic solution.
\newblock International Journal of Approximate Reasoning \textbf{85}, 159--177
  (2017)

\bibitem{Sellamanickam2011APR}
Sellamanickam, S., Garg, P., Keerthi, S.S.: A pairwise ranking based approach
  to learning with positive and unlabeled examples.
\newblock In: Proceedings of the 2011 ACM on Conference on Information and
  Knowledge Management (2011)

\bibitem{shao2015laplacian}
Shao, Y.H., Chen, W.J., Liu, L.M., Deng, N.Y.: Laplacian unit-hyperplane
  learning from positive and unlabeled examples.
\newblock Information Sciences \textbf{314}, 152--168 (2015)

\bibitem{Smola2009RelativeND}
Smola, A.J., Song, L., Teo, C.H.: Relative novelty detection.
\newblock In: The 12th International Conference on Artificial Intelligence and
  Statistics, pp. 536--543 (2009)

\bibitem{srinivasan2001aleph}
Srinivasan, A.: The {A}leph manual (2001)

\bibitem{steinberg1992estimating}
Steinberg, D., Scott~Cardell, N.: Estimating logistic regression models when
  the dependent variable has no variance.
\newblock Communications in Statistics-Theory and Methods \textbf{21}(2),
  423--450 (1992)

\bibitem{Su2006FullBN}
Su, J., Zhang, H.: Full bayesian network classifiers.
\newblock In: Proceedings of the 23rd international conference on Machine
  learning, pp. 897--904. ACM (2006)

\bibitem{Suykens1999LeastSS}
Suykens, J.A.K., Vandewalle, J.: Least squares support vector machine
  classifiers.
\newblock Neural Processing Letters \textbf{9}, 293--300 (1999)

\bibitem{vercrussyen2020}
Vercruyssen, V., Meert, W., Davis, J.: ``now you see it, now you don't!
  ''detecting suspicious pattern absences in continuous time series.
\newblock In: Proceedings of the 2020 SIAM International Conference on Data
  Mining (2020)

\bibitem{vercruyssen2018semi}
Vercruyssen, V., Wannes, M., Gust, V., Koen, M., Ruben, B., Jesse, D.:
  Semi-supervised anomaly detection with an application to water analytics.
\newblock In: Proceedings/IEEE International Conference on Data Mining. IEEE
  (2018)

\bibitem{ward2009presence}
Ward, G., Hastie, T., Barry, S., Elith, J., Leathwick, J.R.: Presence-only data
  and the em algorithm.
\newblock Biometrics \textbf{65}(2), 554--563 (2009)

\bibitem{Webb2005NotSN}
Webb, G.I., Boughton, J.R., Wang, Z.: Not so naive {B}ayes: Aggregating
  one-dependence estimators.
\newblock Machine Learning \textbf{58}, 5--24 (2005)

\bibitem{Xu2014LearningWP}
Xu, Z., Qi, Z., Zhang, J.: Learning with positive and unlabeled examples using
  biased twin support vector machine.
\newblock Neural Computing and Applications \textbf{25}, 1303--1311 (2014)

\bibitem{Yang2014EnsemblePU}
Yang, P., Li, X., Chua, H.N., Kwoh, C.K., Ng, S.K.: Ensemble positive unlabeled
  learning for disease gene identification.
\newblock In: PloS one (2014)

\bibitem{Yang2012PositiveunlabeledLF}
Yang, P., Li, X., Mei, J.P., Kwoh, C.K., Ng, S.K.: Positive-unlabeled learning
  for disease gene identification.
\newblock In: Bioinformatics, pp. 2640--2647 (2012)

\bibitem{Yi2017ScalableDR}
Yi, J., Hsieh, C.J., Varshney, K.R., Zhang, L., Li, Y.: Scalable demand-aware
  recommendation.
\newblock In: Advances in Neural Information Processing Systems, pp. 2412--2421
  (2017)

\bibitem{yu2005single}
Yu, H.: Single-class classification with mapping convergence.
\newblock Machine Learning \textbf{61}(1-3), 49--69 (2005)

\bibitem{Yu2004PEBLWP}
Yu, H., Han, J., Chang, K.C.: {PEBL}: Web page classification without negative
  examples.
\newblock IEEE Transactions on Knowledge and Data Engineering \textbf{16}(1),
  70--81 (2004)

\bibitem{Yu2002PEBLPE}
Yu, H., Han, J., Chang, K.C.C.: {PEBL}: positive example based learning for web
  page classification using svm.
\newblock In: Proceedings of the eighth ACM SIGKDD international conference on
  Knowledge Discovery and Data Mining, pp. 239--248. ACM (2002)

\bibitem{Yu2007PEPUCAG}
Yu, S., Li, C.: Pe-puc: A graph based pu-learning approach for text
  classification.
\newblock In: International Workshop on Machine Learning and Data Mining in
  Pattern Recognition, pp. 574--584. Springer (2007)

\bibitem{Zadrozny2002TransformingCS}
Zadrozny, B., Elkan, C.: Transforming classifier scores into accurate
  multiclass probability estimates.
\newblock In: Proceedings of the eighth ACM SIGKDD international conference on
  Knowledge discovery and data mining, pp. 694--699. ACM (2002)

\bibitem{zhang2009reliable}
Zhang, B., Zuo, W.: Reliable negative extracting based on knn for learning from
  positive and unlabeled examples.
\newblock Journal of Computers \textbf{4}(1), 94--101 (2009)

\bibitem{zhang2005simple}
Zhang, D., Lee, W.S.: A simple probabilistic approach to learning from positive
  and unlabeled examples.
\newblock In: Proceedings of the fifth Annual UK Workshop on Computational
  Intelligence (UKCI), pp. 83--87 (2005)

\bibitem{Zhang2014NonparallelHS}
Zhang, Y., Ju, X., Tian, Y.: Nonparallel hyperplane support vector machine for
  pu learning.
\newblock 2014 10th International Conference on Natural Computation (ICNC) pp.
  703--708 (2014)

\bibitem{Zhao2016Protein}
Zhao, J., Liang, X., Wang, Y., Xu, Z., Liu, Y.: Protein complexes prediction
  via positive and unlabeled learning of the ppi networks.
\newblock In: Proceedings of the 13th International Conference on Service
  Systems and Service Management (ICSSSM), pp. 1--6 (2016).
\newblock \doi{10.1109/ICSSSM.2016.7538432}

\bibitem{zhou2004learning}
Zhou, D., Bousquet, O., Lal, T.N., Weston, J., Sch{\"o}lkopf, B.: Learning with
  local and global consistency.
\newblock In: Advances in Neural Information Processing Systems 17, pp.
  321--328 (2004)

\bibitem{Zhou2012MultiviewPA}
Zhou, J.T., Pan, S.J., Mao, Q., Tsang, I.W.: Multi-view positive and unlabeled
  learning.
\newblock In: Proceedings of the 4th Asian Conference on Machine Learning
  (2012)

\bibitem{Zhou2010LearningWP}
Zhou, K., Xue, G.R., Yang, Q., Yu, Y.: Learning with positive and unlabeled
  examples using topic-sensitive plsa.
\newblock IEEE Transactions on Knowledge and Data Engineering \textbf{22},
  46--58 (2010)

\bibitem{zupanc2018estimating}
Zupanc, K., Davis, J.: Estimating rule quality for knowledge base completion
  with the relationship between coverage assumption.
\newblock In: Proceedings of the Web Conference 2018, pp. 1--9 (2018)

\end{thebibliography}

\end{document}